\documentclass[11pt]{article}

\usepackage[preprint]{acl}

\usepackage{times}
\usepackage{latexsym}

\usepackage{minted}
\usepackage{listings}
\usepackage{pythonhighlight}

\usepackage[T1]{fontenc}

\usepackage[utf8]{inputenc}

\usepackage{microtype}

\usepackage{inconsolata}

\usepackage{graphicx}
\usepackage{amsmath}
\usepackage{enumitem}
\usepackage{graphicx}
\usepackage{booktabs}
\usepackage{multirow}
\usepackage{amssymb}
\usepackage{bm}

\title{Reinforcement Learning for Self-Improving Agent with Skill Library}

\author{Jiongxiao Wang$^{1}$\thanks{Work was done during an internship at AWS Agentic AI.} \quad Qiaojing Yan$^{2}$ \quad Yawei Wang$^{2}$ \quad Yijun Tian$^{2}$ \quad Soumya Smruti Mishra$^{2}$ \\ \textbf{Zhichao Xu}$^{2}$ \quad  \textbf{Megha Gandhi}$^{2}$ \quad \textbf{Panpan Xu}$^{2}$\thanks{Corresponding author: Panpan Xu, Email: xupanpan@\\amazon.com} \quad \textbf{Lin Lee Cheong}$^{2}$ \\
\textsuperscript{1}University of Wisconsin–Madison; \textsuperscript{2}AWS Agentic AI \\ 
\small{
\texttt{jwang2929@wisc.edu; \{qiaojiny, yawenwan, yijunt, soumish, xzhichao, ganmegha, xupanpan, lcheong\}@amazon.com}}
}

\newcommand{\name}{SAGE}
\newcommand{\fullname}{Skill Augmented GRPO for self-Evolution}

\begin{document}
\maketitle
\begin{abstract}
Large Language Model (LLM)-based agents have demonstrated remarkable capabilities in complex reasoning and multi-turn interactions but struggle to continuously improve and adapt when deployed in new environments. One promising approach is implementing skill libraries that allow agents to learn, validate, and apply new skills. However, current skill library approaches rely primarily on LLM prompting, making consistent skill library implementation challenging.
To overcome these challenges, we propose a Reinforcement Learning (RL)-based approach to enhance agents' self-improvement capabilities with a skill library. Specifically, we introduce \fullname{} (\name), a novel RL framework that systematically incorporates skills into learning. The framework's key component, Sequential Rollout, iteratively deploys agents across a chain of similar tasks for each rollout. As agents navigate through the task chain, skills generated from previous tasks accumulate in the library and become available for subsequent tasks. Additionally, the framework enhances skill generation and utilization through a Skill-integrated Reward that complements the original outcome-based rewards.
Experimental results on AppWorld demonstrate that \name{}, when applied to supervised-finetuned model with expert experience, achieves 8.9\% higher Scenario Goal Completion while requiring 26\% fewer interaction steps and generating 59\% fewer tokens, substantially outperforming existing approaches in both accuracy and efficiency. Our code is available at \href{https://github.com/amazon-science/SAGE}{https://github.com/amazon-science/SAGE}.
\end{abstract}

\section{Introduction}
Large language model (LLM)-based agents have been widely applied to automate complex tasks through active environmental interactions, including coding agent \citep{yang2024swe, novikov2025alphaevolve}, deep research \citep{deep_research}, assistant agent \cite{yao2024tau, chen2025reinforcement},
and web browsing \citep{yao2022webshop, zhouwebarena}. To enhance the performance of these multi-turn interactive agents, researchers have successfully integrated reinforcement learning (RL) techniques into their frameworks \citep{chen2025reinforcement, qiwebrl, zhou2025sweet, wang2025ragen}. Recent advances, particularly in reinforcement learning with verifiable rewards (RLVR) \citep{shao2024deepseekmath, guo2025deepseek}, have enabled effective end-to-end agent training for improved performance \citep{jin2025search}. However, despite RL's effectiveness, significant limitations persist: RL-trained agents are often limited to specific training scenarios \citep{zheng2024gpt, li2024effects}. When deployed in new environments, they struggle to demonstrate continual learning capabilities to effectively utilize valuable on-going experiences for future tasks.

To address these limitations, one potential solution is enabling agents to transform their previous interaction experiences into reusable skills, which can be stored in a skill library for future reference. When agents encounter similar tasks, these previously acquired skills can be leveraged through experience replay to improve task success rates, particularly in scenarios that were not encountered during training but experienced during deployment. Furthermore, since each skill is composed of a list of actions, utilization of these skills can enhance agent efficiency by condensing complex action sequences into reusable operations.

\begin{figure*}[ht]
    \centering
    \includegraphics[width=1.0\textwidth]{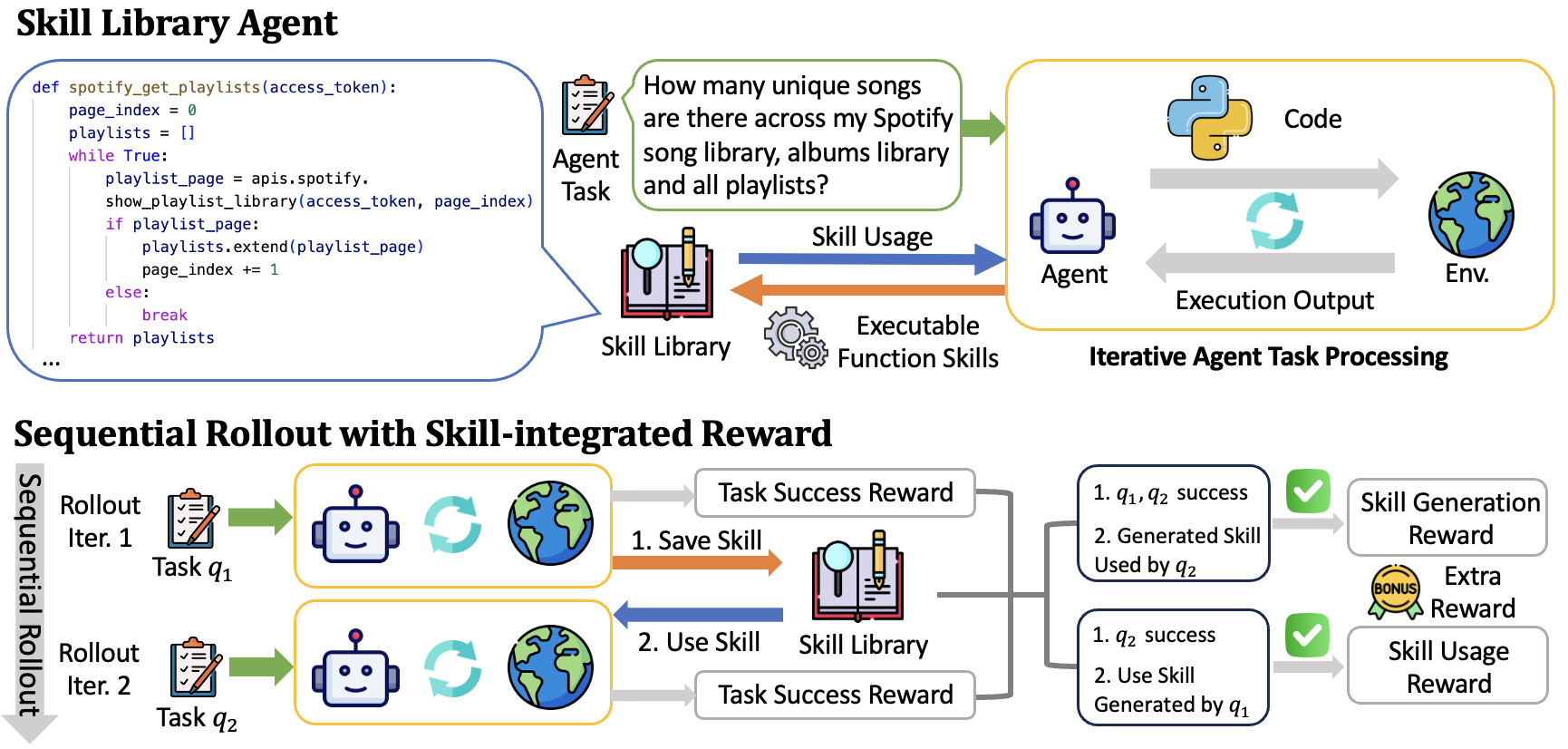}
    \caption{Illustration for Skill Library Agent and Sequential Rollout with Skill-integrated Reward.}
    \label{fig:intro}
\end{figure*}

Recent research has made significant strides in enabling agents to compose reusable skills \citep{wangvoyager, cailarge, nguyen2024dynasaur, wang2024agent, zheng2025skillweaver, wang2025inducing}. For instance, \citet{wang2025inducing} focuses on inducing high-level web browsing skills from successful action trajectories, transforming primitive actions like \textit{click} and \textit{search} into more complex operations such as \textit{search product}. While these works have demonstrated the effectiveness of skill libraries for deployed agents, they predominantly rely on manually crafted prompts for skill generation and utilization. This prompt-based approach, however, is inherently constrained by the instruction following capabilities of the base model, limiting both the quality and adaptability of the skill library.

In this paper, we explore enhancing agents' self-improvement capabilities through RL with a skill library. Given the diverse frameworks of agents for various tasks, we focus specifically on tool-using agents that interact with environments through API calls and receive corresponding feedback. Our implementation extends the CodeAct framework \citep{wang2024executable}, which enables agents to compose code by combining multiple APIs with basic programming constructs (such as for loops) to solve complex tasks. Building upon this foundation, we develop a specialized framework for self-improving agents with skill library (referred to as \textbf{skill library agents} for simplicity). Unlike previous frameworks such as Agent Skill Induction \citep{wang2025inducing}, which typically define reusable skills only after task completion, we implement a unified format for both task solving and skill generation, following \citet{nguyen2024dynasaur}. In our approach, when the agent model interacts with the environment through APIs, it generates programmatic functions that can be saved as skills and subsequently called to execute, rather than using multiple APIs directly. Given that in-context prompting struggles to adapt open-source models to this newly proposed skill library agent, supervised fine-tuning (SFT) is first applied before RL using high-quality trajectories collected through expert experience generated by advanced LLMs.

Based on the SFT model, we propose a novel RL framework for skill library agents. Traditional RL approaches typically consider rewards only for individual examples, limiting their scope to ongoing task performance. 
For skill library agents, the focus is on developing high-quality, reusable skills while improving accurate skill usage from the library for task optimization.
To realize this, we extend Group Relative Policy Optimization (GRPO) \citep{shao2024deepseekmath} and propose \fullname{} (\name), consisting of \textbf{Sequential Rollout} and \textbf{Skill-integrated Reward}. Instead of single-task trajectories, the agent is trained with chains of similar tasks. We implement the Sequential Rollout process on the task chain, where skills generated in the previous tasks are preserved and made available for use in the following ones. Under this framework, the final Skill-integrated Reward is computed as the sum of two components: the verifiable outcome-based reward and an extra reward for high-quality skill generation and utilization.

To evaluate the effectiveness of \name{} for skill library agents, we conducted experiments on the AppWorld dataset \citep{trivedi2024appworld}, where agents interact with their environment through API documentation lookup, API calls, and logical programming constructs to solve complex practical tasks. AppWorld utilizes the Scenario Goal Completion (SGC) metric, which measures the success rate of scenarios where all three similar tasks within a scenario are successfully completed. This metric effectively evaluates how well generated skills transfer across similar tasks. After applying \name{} to Qwen2.5-32B-Instruct, we observe significant improvements compared to prompting-based approaches. Our method achieves more than 3x SGC score while requiring less than half of generated tokens. Analysis of skill utilization reveals that agents trained with \name{} demonstrate more than 2x success rates when utilizing learned skills.

Furthermore, our approach achieves state-of-the-art performance on both Test Normal and Test Challenge datasets compared to previous training methods which also applied RL but without a skill library. For example, our method achieves 72.0\% Task Goal Completion (TGC) and 60.7\% SGC with an average of 12.1 interaction steps and 1,475 generated tokens on the Test Normal set. This represents substantial improvement over baseline training with GRPO, which achieved 69.2\% TGC and 51.8\% SGC while requiring 16.4 average steps and 3,613 tokens. These improvements demonstrate the effectiveness of our approach in enhancing both task performance and efficiency through the skill library agent.

\section{Related Work}

\noindent \textbf{LLM-based Agent.} Recent advancements in instruction-tuned LLMs have enabled them to follow user instructions for interacting with external environments as autonomous agents. Various frameworks have been developed to enhance these backbone LLMs' capabilities in performing agentic tasks. \citet{yao2023react} pioneered a "reason-then-act" pipeline, guiding LLMs to generate interactive actions for agents after a text reasoning process. \citet{erdoganplan} extended this approach by incorporating an additional planning phase. Additionally, \citet{wang2024executable} demonstrated that generating executable Python code and then run it within a code interpreter could significantly improve agent performance. To further enhance agent capabilities, researchers have applied both supervised fine-tuning \citep{schick2023toolformer, chen2023fireact, zeng2024agenttuning, zhang2024agentohana} and reinforcement learning \citep{song2024trial, bai2024digirl, wang2025ragen, chen2025reinforcement, qiwebrl, zhou2025sweet} to develop more effective and adaptive agent backbone LLMs.

\noindent \textbf{Self-Improving Agent with Skill Library.} While RL algorithms have enabled agents to self-improve through valuable experiences explored during rollouts \citep{zhouproposer, putta2024agent, qiwebrl}, enabling continuous self-improvement after deployment, especially in new environments, remains challenging. \citet{wangvoyager} pioneered the use of a skill library to record successful behaviors for later retrieval in Minecraft exploration. Subsequently, numerous studies have demonstrated the effectiveness of such skill libraries across various agentic tasks, including web exploration \citep{wang2024agent, zheng2025skillweaver, wang2025inducing}, computer control \citep{zhengsynapse, wu2024copilot}, and math problems \citep{nguyen2024dynasaur}. These skills can take two forms: natural language experience memories serving as a reference, or executable skills that can be directly implemented in the environment. In this paper, we focus on leveraging RL to enhance agents' self-improvement capabilities by teaching them to generate executable skills for the skill library during test time.

\section{Method}
In this section, we first detail the skill library agent. We then present our new RL framework specifically designed to integrate the skill library during training process. An illustrative figure of our skill library agent and Sequential Rollout with Skill-integrated Reward design is shown in Figure~\ref{fig:intro}.

\subsection{Skill Library Agent}
Before implementing RL, we first need to develop a self-improving agent that integrates the skill library for tool-using agents. This framework will serve as the foundation for RL rollout process and task evaluations.

Existing frameworks for skill library agents, such as Agent Skill Induction \citep{wang2025inducing}, Agent Workflow Memory \citep{wang2024agent}, and Voyager \citep{wangvoyager}, typically define reusable skills after completing entire tasks. While this approach allows agents to observe complete task trajectories before determining skill definitions, it limits the RL process in two ways: (1) In long-horizon tasks, the additional skill generation process further extends the context length, potentially exceeding the model's limitations; (2) The separation between task execution and skill generation creates an inconsistency that may impact learning effectiveness.

To address these limitations, we follow the DynaSaur approach \cite{nguyen2024dynasaur} and implement a unified format for both task solving and skill generation. Specifically, when the agent model interacts with the API environments, it first generates a skill function and then calls it to process, instead of using multiple APIs. An example of the format difference is presented in Appendix~\ref{appendix:format}.

Formally, given a task set $Q$, our agent is designed to perform online learning from start to finish with a skill library $\mathcal{M}$, which can be initialized with either an empty set or previously defined skills. For each task $q\sim Q$, the agent first retrieves a skill subset $[a_1,...,a_k]$ from the skill library $\mathcal{M}$ and adds them into context before performing the task. After that, the agent can automatically perform the following actions to interact with the skill library: (1) \textbf{Skill Usage}: Perform skill $a_i \in [a_1,...,a_k]$ to process the task; (2) \textbf{Skill Generation}: Define a skill function $\hat{a}$ composed of multiple actions aimed at solving the task and then immediately call it to process the task; (3) \textbf{Skill Update}: If the skill $a$, either from the skill library or newly defined, fails to execute, update the skill and recall it to process the task; (4) \textbf{Skill Save}:  If the skill can be executed without error, add the new skill or update the existing skill in the skill library $\mathcal{M}$. Additionally, direct API calls are allowed for cases where defining a function skill is unnecessary for task processing.


\subsection{\name{} for Skill Library Agent}
We begin the description of our \name{} framework by introducing the preliminary GRPO algorithms, followed by our specifically-designed components for skill library agents: Sequential Rollout and Skill-integrated Reward.

\subsubsection{Preliminary}
In this paper, we build our \name{} based on GRPO \citep{shao2024deepseekmath}. For each query $q$, GRPO first samples a group of outputs $\{o_1, ..., o_G\}$ from the old policy $\pi_{\theta_{old}}$ and then optimizes the policy by maximizing the following objective:
\begin{align}
\mathcal{J}_{\text{GRPO}}(\theta) =& \mathbb{E}_{[q \sim Q, \{o_i\}_{i=1}^G \sim \pi_{\theta_{old}}(O|q)]} \nonumber \\
& \frac{1}{G}\sum_{i=1}^G\frac{1}{|o_i|}\sum_{t=1}^{|o_i|} 
\{\min[s_{i,t}\hat{A}_{i,t}, \nonumber \\
&\text{clip}\left(s_{i,t},1-\epsilon, 1+\epsilon\right)\hat{A}_{i,t}] -\beta\mathbb{D}_{\text{KL}} \} \nonumber
\end{align}
\noindent where $s_{i,t} = \frac{\pi_\theta(o_{i,t}|q,o_{i,<t})}{\pi_{\theta_{old}}(o_{i,t}|q,o_{i,<t})}$, $\epsilon$ is the clip ratio and $\beta$ is the ratio of KL penalty between policy model $\pi_\theta$ and reference model $\pi_{ref}$. $\hat{A}_{i,t}$ is the advantage calculated based on the relative rewards of the outputs inside each group. Given rewards $\mathbf{r} = \{r_1, r_2, ..., r_G\}$ of the outputs under the same group, the advantage is defined as $\hat{A}_{i,t}=\frac{r_i-\text{mean}(\mathbf{r})}{\text{std}(\mathbf{r})}$.

\subsubsection{Sequential Rollout}
A principled method to augment agents' self-improving ability with skill library is through end-to-end RL. However, skill generation and usage processes often need multiple tasks to reveal the quality of skills. One potential solution to enabling the end-to-end RL is that instead of one task, we could give the agent a chain of tasks. Then a sequential rollout process is performed across these tasks, enabling the agent to progressively accumulate skills in its library throughout the task chain. Any skills learned in earlier tasks could be used in subsequent tasks. In this way, rewards signal from the successful usage of skills in later tasks can be back-propagated to the skill generation in the previous tasks. To ensure that the generated skills can be immediately applied to subsequent tasks, we construct the sequential tasks with examples under the same scenario, in which all tasks share similar instructions.

While a longer task chain with multiple examples would better approximate practical sequential evaluation processes on the whole test set, it would significantly increase training costs. For simplicity, this paper focuses on task chains containing only two examples. A detailed discussion of Sequential Rollout with extended task chains is provided in Appendix~\ref{appendix:long_chain}.


\subsubsection{Skill-integrated Reward}
Unlike baseline RL that relies on outcome-based rewards, our training framework incorporates additional rewards specifically designed for the agent's interactions with the skill library. For this purpose, we introduce the Skill-integrated Reward. Specifically, we aim to encourage two additional behaviors across a two-example task chain: skill generation in the first example and skill utilization in the second example.

Formally, let $ r^1, r^2 \in [0,1]$ denote the verifiable outcome-based rewards of the task chain $(q^1, q^2)$ collected from Sequential Rollout, where the skills generated by $q^1$ are directly utilized by $q^2$. We formulate Skill-integrated Rewards $R^1$ and $R^2$ as:
\begin{equation*}
\begin{cases}
R^1 =r^1 + \mathbf{1}[r^1 = 1] * \mathbf{1}[r^2 = 1] * \mathbf{1}_{skill}(q^2|q^1) \\
R^2 =r^2 + \mathbf{1}[r^2 = 1] * \mathbf{1}_{skill}(q^2|q^1)
\end{cases}
\end{equation*}
\noindent where indicator $\mathbf{1}_{skill}(q^2|q^1)$ denotes whether $q^2$ uses skills generated by $q^1$; and $\mathbf{1}[r = 1]$ represents whether the outcome-based reward equals 1, i.e. successful task completion. To ensure the skill library agent adheres to a specific format that generates code for each interaction, we impose a -1.0 penalty reward specifically on responses where the agent provides no code and terminates the task.


\subsubsection{\name}

After collecting the rollout trajectories and computing the corresponding rewards, we implement our \fullname{} (\name{}) to train the skill library agent. Following \citet{chen2025reinforcement}, we choose not to use the KL divergence penalty and the advantage would not be normalized by standard deviation of the rewards.

Given the current policy $\pi_\theta$ and the old policy $\pi_{\theta_{old}}$, we first sample task chains from the original data distribution: $(q^1, q^2) \sim Q$. For each task pair, we then perform $G$ group rollout $\{\tau_i\}_{i=1}^G \sim \pi_{\theta_\text{old}}(\cdot|(q^1,q^2))$, where $\tau_i$ represents the Sequential Rollout trajectory collected by sequentially processing $(q^1,q^2)$. To differentiate outputs for each task query in $\tau_i$, we define $o_i^k \in \tau_i$ as the $i^\text{th}$ output for $q^k$ in the group, where $k$ is the chain index. The current policy is then optimized by maximizing the following objective function for skill library-integrated GRPO: 
\begin{align}
\mathcal{J}_{Agent}(\theta) &= \mathbb{E}_{{\color{red}(q^1, q^2) \sim Q, \{\tau_i\}_{i=1}^G \sim \pi_{\theta_\text{old}}(\cdot|(q^1, q^2))}} \nonumber \\
&\frac{1}{G}\sum_{i=1}^G {\color{red}\sum_{k=1}^2} \frac{1}{|o_i^k|}\sum_{t=1}^{|o_i^k|}\{\min[s_{i,t}^k\hat{A}_{i}^k, \nonumber \\
&\text{clip}\left(s_{i,t}^k,1-\epsilon, 1+\epsilon\right)\hat{A}_{i}^k]\} \nonumber
\end{align}
\noindent where $s_{i,t}^k = \frac{\pi_\theta(o_{i,t}^k|q^k,{\color{red}\mathcal{M}_{i}^k},o_{i,<t}^k)}{\pi_{\theta_{old}}(o_{i,t}^k|q^k,{\color{red}\mathcal{M}_{i}^k},o_{i,<t}^k))}$; and $\hat{A}_{i}^k=R_{i}^k-\text{mean}(\{R_{i}^k|i=1,2,..,G\})$. $s_{i,t}^k$ represents the importance sampling term; $\hat{A}_{i}^k$ is the advantage computed by the Skill-integrated Reward $R_{i}^k$; $\mathcal{M}_{i}^k$ denotes the skill library integrated with the query $q_i^k$ for group $i$. Here $\mathcal{M}_{i}^1$ is an empty set and $\mathcal{M}_{i}^2$ includes skills generated when performing $q_i^1$. Since our skill library agents need multi-turn interactions with the environments to process the task, the outputs $|o_{i}^k|$ only consider the LLM generation contents, while the observations from the environment are masked.

The key differences between our approach and the original GRPO described in Section 3.2.1 are highlighted in red. In our method, due to the Sequential Rollout mechanism, the expectation is computed across the task chain. Notably, within the same group, the generations $o_{i}^2$ are derived from different skill libraries $\mathcal{M}_{i}^2$, unlike the original GRPO where generations stem from identical queries.
\section{Experiments}
We begin this section with the details of our experimental settings followed by the presentation of our main results. Further analysis and ablation studies are conducted to better demonstrate the effectiveness of \name.

\subsection{Experimental Settings}

\noindent \textbf{Dataset and Base Model.} To evaluate the effectiveness of \name{} for skill library agents, particularly for those designed for long-horizon tool usage, we adopt the AppWorld dataset \citep{trivedi2024appworld}. AppWorld contains 750 tasks from 250 task scenarios. Each scenario consists of three tasks sharing similar instructions. The scenario-based structure of AppWorld makes it well-suited for Sequential Rollout, as tasks within the same scenario naturally form a task chain. We utilize the Train set for all training steps, including SFT and subsequent \name{}. The Dev set guides the best checkpoint selection during training. Evaluation is performed on both Test-Normal and Test-Challenge splits. More details about the dataset are presented in Appendix~\ref{appendix:dataset}.

To ensure a fair comparison with the previous work \citep{chen2025reinforcement} on AppWorld, we used Qwen2.5-32B-Instruct \citep{qwen2.5} as our base model for training purpose.

\noindent \textbf{Skill Library Agent.} Following the ReAct agent \citep{yao2023react} as presented in the AppWorld paper \citep{trivedi2024appworld}, we implemented our skill library agent with a specifically designed in-context example and detailed instructions to guide the base model in using the skill library to perform complex tasks. The detailed prompt of our skill library agent is presented in Appendix~\ref{appendix:prompt}. Regarding skill retrieval, we employed an idealized case during sequential rollout where skills are retained and utilized only within the same scenario in the task chain. This approach eliminates the need for a retrieval model, significantly simplifying the RL rollout processes. While our default evaluation uses the ideal case (Same Scenario) with provided scenario labels from the test set, we conducted additional analysis on various retrieval methods to address real-world scenarios where such labels may not be available.


\noindent \textbf{Baseline GRPO.} Since the baseline method LOOP \citep{chen2025reinforcement} does not provide source code or model, and their paper does not report agent efficiency metrics, it is challenging to make a direct, fair comparison. Therefore, we implemented our own baseline training using the GRPO method without Skill Library for comparison purposes. Following \citet{chen2025reinforcement}, our baseline GRPO removes the KL divergence penalty and calculates advantage using the mean reward within the group instead of normalizing by the standard error. The only difference between our baseline GRPO and LOOP is that we perform strictly on-policy RL without PPO epochs. More training settings of Baseline GRPO are presented in Appendix~\ref{appendix:grpo}. 

\noindent \textbf{\name{}.} Initial RL experiments with open-source models revealed their limited capabilities in following instructions when integrated with our skill library agent, despite carefully designed prompts. This limitation resulted in insufficient self-improvement capabilities, hampering the generation of high-quality sequential rollouts necessary for our \name{}. Thus, we first implemented supervised fine-tuning prior to RL using an expert experience dataset. For simplicity, we employed an advanced model, specifically Claude 3.5 Sonnet V2, as the expert to produce high-quality trajectories within the skill library agent. The experimental details of expert data generation for SFT are shown in Appendix~\ref{appendix:sft}. Based on the SFT model, we then performed our \name{} with sequential rollout and skill-integrated reward. Details of training parameters and processes are listed in Appendix~\ref{appendix:sgrpo}.


%




\noindent \textbf{Metrics and Evaluation.} 
We assessed performance using two primary metrics: \textbf{Task Goal Completion (TGC)} measures the accuracy of successfully completed individual tasks; and \textbf{Scenario Goal Completion (SGC)} calculates the proportion of scenarios where all three included tasks are successfully completed. To evaluate efficiency, we counted average interaction steps (\textbf{Avg. Steps}) and average generated tokens (\textbf{Avg. Tokens}) required for task completion, where fewer steps and tokens indicate more efficiency. All reported numbers in this paper represent the average of three agent evaluation runs using the same model.

\begin{table*}[ht]
\caption{Task Performance of \name{} compared with various baselines. Results of Methods marked with "*" are taken from \citep{chen2025reinforcement}. In these cases, Avg. Steps and Tokens were not reported, thus denoted by "- -".}
\label{table1}
\begin{center}
\resizebox{2.1\columnwidth}{!}{
\begin{tabular}{cccccccccc}
\toprule
\multirow{2}{*}{Base Model} & \multirow{2}{*}{Methods} & \multicolumn{4}{c}{Test Normal} & \multicolumn{4}{c}{Test Challenge}  \\
 & & TGC & SGC& Avg. Steps & Avg. Tokens & TGC & SGC& Avg. Steps &Avg. Token \\
\midrule
\multicolumn{10}{c}{\textbf{Training Free Methods}} \\
\midrule
GPT-4o & ReAct* & 48.8 & 32.1 & - - & - - & 30.2 & 13.0 & - - & - -\\
OpenAI o1 & ReAct* & 61.9 & 41.1 & - - & - - & 36.7 & 19.4 & - - & - -\\
Claude Sonnet 3.5 V2 & ReAct & 57.1 & 41.1 & 15.7& 1,542 & 49.2 & 28.8 & 21.8& 2,084\\
Qwen2.5 32B Instruct & ReAct* & 39.2 ± 3.5 & 18.6 ± 2.0 & - -& - - & 21.0 ± 1.4 & 7.5 ± 1.2 & - - & - -\\ 
\midrule
\multicolumn{10}{c}{\textbf{RL without Skill Library}} \\
\midrule
\multirow{2}{*}{Qwen2.5 32B Instruct} & LOOP* &71.3 ± 1.3 &  53.6 ± 2.2 & - - & - -&45.7 ± 1.3 & 26.6 ± 1.5 & - -& - -\\ 
 & GRPO & 69.2 ± 2.7 & 51.8 ± 5.8 & 16.4 ± 0.2& 3,613 ± 200& 40.7 ± 1.5 & 26.9 ± 1.5& 21.9 ± 0.1& 5,211 ± 65\\
\midrule
\multicolumn{10}{c}{\textbf{Our Approach}} \\
\midrule
\multirow{3}{*}{Qwen2.5 32B Instruct} & \multicolumn{1}{l}{Skill Library Agent} & 30.7 ± 3.1 & 19.6 ± 1.4 & 13.4 ± 0.4& 2,988 ± 73 & 15.3 ± 1.7 & 7.0 ± 1.2 & 18.7 ± 0.4 & 4,803 ± 117\\ 

 & \multicolumn{1}{l}{{ \ + SFT}} & 55.2 ± 1.5 & 41.7 ± 1.7 & \textbf{11.4 ± 0.5} & \textbf{1,340 ± 65} &37.2 ± 1.2 & 20.9 ± 1.8 & \textbf{16.2 ± 0.3} & 1,909 ± 80\\
 
 & \multicolumn{1}{l}{{ \ + \name{}}} & \textbf{72.0 ± 1.5} & \textbf{60.7 ± 1.5} & 12.1 ± 0.2 & 1,475 ± 127 &\textbf{50.1 ± 2.0} & \textbf{32.4 ± 3.7} & 17.3 ± 0.3 & \textbf{1,807 ± 29}\\ 
\bottomrule
\end{tabular}
}
\end{center}
\end{table*}

\subsection{Main Results}
Table~\ref{table1} presents our main results of \name{} compared with various baselines. From the table, we can observe that \name{} demonstrates significantly improved performance compared to the baseline GRPO, particularly a 8.9\% improvement in SGC (Scenario Goal Completion) on test normal dataset. Since SGC measures the agent's performance across multiple related tasks within a scenario, this improvement demonstrates our model's ability to effectively transfer and reuse skills across similar tasks. Additionally, skill library agent trained by \name{} significantly reduces average interaction steps and generated tokens, with 59\% less tokens compared to the baseline agent trained by GRPO. This demonstrates that skill reuse can accomplish complex tasks more efficiently at lower cost. Although our final results rely on SFT using expert experience data generated by Claude, our RL approach with the skill library enables open-source models to surpass the expert performance.

For the results of our approach, we presented the findings in a stepwise manner to better demonstrate the improvements achieved at each stage (prompting based skill library agent, SFT, and \name{}). Initially, when compared to the baseline training-free approach using the Qwen2.5 32B Instruction model with ReAct agent, our skill library agent shows lower performance, indicating the limitations of prompt-based methods for self-improving agent with skill library. The performance significantly improves after SFT with expert experience generated by Claude, though still not surpassing the baseline GRPO without skill library. This indicates that merely imitating expert behaviors is insufficient for optimized performance. Ultimately, our RL method further enhances the SFT-trained model, achieving superior performance compared to all baselines. To better understand how our approach enhances task performance with skill library, we provide examples of actual task execution across different agent models in Appendix~\ref{appendix:example}.


\subsection{Skill Library Usage Analysis}

\begin{figure}
    \centering
    \includegraphics[width=0.5\textwidth]{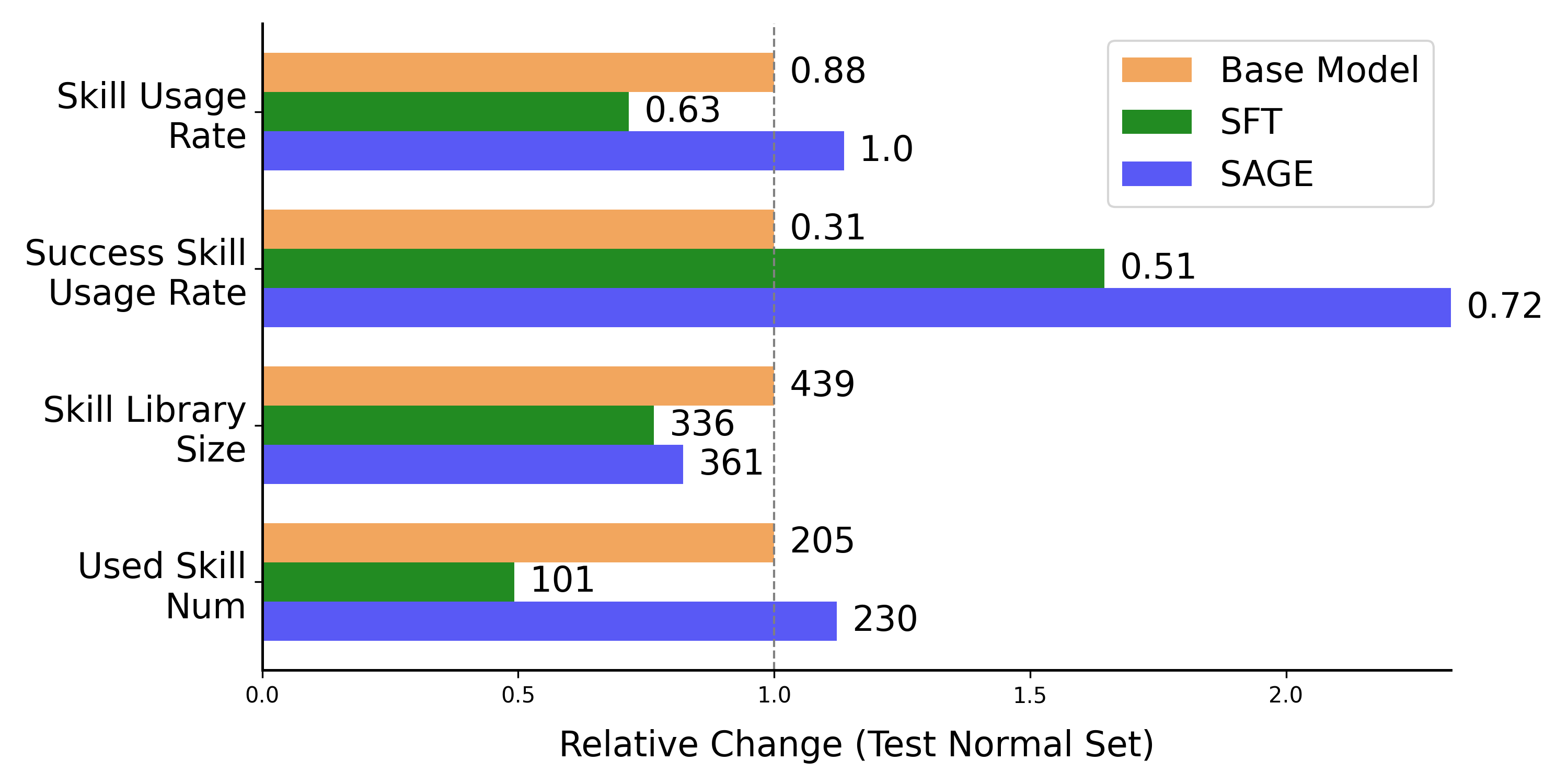}
    \caption{Analysis of Skill Usage Patterns. Performance metrics are shown as ratios relative to Base Model baseline, with numerical values annotated.}
    \label{fig:analysis}
\end{figure}

Though \name{} has reached state-of-the-art performance on AppWorld, it remains unclear how skill library are actually involved during evaluation. This section provides a detailed analysis of skill library usage patterns at each stage of our approach. We employ various metrics to evaluate skill library utilization during evaluation. (1) \textbf{Skill Usage Rate}: Among examples with skill library, the proportion that use skills; (2) \textbf{Success Skill Usage Rate}: Among examples that use skills, the proportion that reach successful task completion; (3) \textbf{Skill Library Size}: Total number of generated skills in the skill library; (4) \textbf{Used Skill Num}: Number of skills in the skill library being used. The analysis results are summarized in Figure~\ref{fig:analysis}.

Analysis of the results reveals that \name{} significantly improves both Skill Usage Rate and Success Skill Usage Rate compared to previous steps. This improvement demonstrates enhanced skill utilization and self-improving capabilities developed during the RL process. The Base Model shows high Skill Usage Rate and Skill Library Size, indicating its fundamental ability to generate and use skills in response to instructional prompts. Although the Base Model generates more skills than \name{}, its lower Used Skill Num and Success Skill Usage Rate indicate limitations in both skill generation quality and utilization effectiveness. Notably, the SFT model only surpasses the Base Model in Success Skill Usage Rate. This outcome may be attributed to the limitations of SFT from expert experiences. While these experiences enhance overall performance, they appear insufficient for developing the model's self-improvement capabilities in skill generation and usage.

\subsection{Ablation Studies}

To further demonstrate the effectiveness of \name{}, we perform ablation studies shown as follows:

\noindent \textbf{Evaluation without Skills.} To validate that the inclusion of the skill library indeed contributes to better performance, we conducted additional evaluations of the skill library agent but with an empty skill library, on the test normal dataset. 
\begin{table}[ht]
\caption{Stepwise performance of skill library agent compared with no skills involved.}
\label{table2}
\centering
\resizebox{\columnwidth}{!}{
\begin{tabular}{cccccc}
\toprule
\multirow{2}{*}{Step} & \multirow{2}{*}{With Skill} & \multicolumn{4}{c}{Test Normal}  \\
 & & TGC & SGC& Avg. Steps & Avg. Tokens  \\
\midrule
Skill Library & \checkmark & 30.7 ± 3.1 & 19.6 ± 1.4  & 13.4 ± 0.4 & 2,988 ± 73  \\ 
Agent& $\bm{\times}$ & 34.7 ± 3.0 & 14.9 ± 3.1  & 16.4 ± 0.5 & 3,704 ± 139\\ 
\midrule
\multirow{2}{*}{SFT} & \checkmark & 55.2 ± 1.5 & 41.7 ± 1.7 & \textbf{11.4 ± 0.5}  & \textbf{1,340 ± 65} \\ 
 & $\bm{\times}$ & 54.8 ± 2.1 & 39.9 ± 2.2 & 13.5 ± 0.1 & 1,611 ± 11 \\ 
\midrule
\multirow{2}{*}{\name{}} & \checkmark & \textbf{72.0 ± 1.5} & \textbf{60.7 ± 1.5} & 12.1 ± 0.2 	& 1,475 ± 127 \\ 
 & $\bm{\times}$ & 71.4 ± 0.5 & 54.8 ± 0.8 & 16.0 ± 0.2 & 1,937 ± 81 \\ 
\bottomrule
\end{tabular}
}
\end{table}

As shown in Table~\ref{table2}, all agent models achieve improved SGC scores with reduced average steps and tokens when employing skills compared to no skill settings. This improvement highlights the importance of incorporating the skill library to enhance SGC performance and task efficiency, even in the absence of training. However, we observe a decline in TGC scores for Skill Library Agent. This decline may be attributed to the models' limited proficiency in skill utilization, resulting in inappropriate skill applications and consequent task failures, as detailed in Section 4.3.


\noindent \textbf{Skill Library Agent with Retrieval in Practice.} 
In practical applications, tasks often lack explicit scenario information that would enable direct skill acquisition from the same scenario as used during training. To address this limitation, we implemented an additional retrieval process that identifies and leverages relevant skills from previous experiences during evaluation. Specifically, we explored three different retrieval methods: Query N-gram, Query Embedding, and Skill Embedding.
\begin{table}[ht]
\caption{\name{} with different retrieval methods.}
\label{table4}
\begin{center}
\resizebox{1.0\columnwidth}{!}{
\begin{tabular}{ccccc}
\toprule
 \multirow{2}{*}{Retrieval Methods}  & \multicolumn{4}{c}{Test Normal}  \\
 & TGC & SGC& Avg. Steps & Avg. Tokens  \\
\midrule
Same Scenario  &\textbf{72.0 ± 1.5}  & \textbf{60.7 ± 1.5}& 12.1 ± 0.2  & 1,475 ± 127 \\ 
Query N-gram  & \textbf{72.0 ± 2.0} & 60.1 ± 1.7 & 12.7 ± 0.3 & 1,466 ± 101 \\ 
Query Embedding & 69.6 ± 1.6 & 59.5 ± 2.2 & \textbf{11.8 ± 0.2} & \textbf{1,335 ± 63} \\ 
Skill Embedding & 66.3 ± 0.7 & 56.0 ± 0.8 & 14.5 ± 0.3 & 1,692 ± 10 \\ 
\bottomrule
\end{tabular}
}
\end{center}
\end{table}

The results in Table~\ref{table4} show that retrieval method selection significantly impacts performance. Carefully designed skill retrieval mechanisms can approach or even partially outperform the best performance achieved under Same Scenario conditions. 
A comprehensive description of each retrieval method, along with detailed results analysis, can be found in Appendix~\ref{appendix:retrieval}.

\noindent \textbf{Reward Design.} To demonstrate the effectiveness of our Skill-integrated Reward design, we conducted ablation studies comparing it with two alternative reward designs: Outcome-based Reward and Chain-based Reward. Details of the alternative reward designs and their corresponding formulations are included in the Appendix~\ref{appendix:reward1}. 
\begin{table}[ht]
\caption{\name{} with different reward designs.}
\label{tbl:reward}
\begin{center}
\resizebox{\columnwidth}{!}{
\begin{tabular}{ccccc}
\toprule
 \multirow{2}{*}{Reward Design} & \multicolumn{4}{c}{Test Normal}  \\
 & TGC & SGC& Avg. Steps & Avg. Tokens \\
\midrule
Skill-integrated & \textbf{72.0 ± 1.5}  & \textbf{60.7 ± 1.5}& \textbf{12.1 ± 0.2} & 1,475 ± 127\\ 
Outcome-based & 69.8 ± 1.0 & 55.4 ± 1.4 & 13.1 ± 0.2 & 1,469 ± 61 \\ 
Chain-based & 67.9 ± 2.1  & 56.6 ± 1.6 & 15.7 ± 0.3 & \textbf{1,361 ± 58}\\ 
\bottomrule
\end{tabular}
}
\end{center}
\end{table}

As shown in Table~\ref{tbl:reward}, our Skill-integrated Reward design achieves superior TGC and SGC scores compared to alternative reward designs. Regarding efficiency, while Skill-integrated Reward yields the lowest Avg. Seps, it results in relatively high Avg. Tokens. This trade-off likely stems from skill usage reducing the number of steps while requiring additional tokens for skill generation. We further analyze the skill library usage behaviors across different reward designs through additional studies (similar to Section 4.3) in Appendix~\ref{appendix:reward2}.


\noindent \textbf{RL Initialization.} In this paper, we initialized \name{} using SFT. To demonstrate the necessity of using extra data for SFT before RL, we conducted ablation studies with various initialization methods including Base Model, Self-Distillation and RL Warm-Up (detailed in Appendix~\ref{appendix:init}).
\begin{table}[ht]
\caption{\name{} initialized with different methods.}
\label{tbl:initialization}
\begin{center}
\resizebox{\columnwidth}{!}{
\begin{tabular}{cccccc}
\toprule
Initialization & Extra & \multicolumn{4}{c}{Test Normal}  \\
 Methods& Data & TGC & SGC& Avg. Steps & Avg. Tokens  \\
\midrule
Base Model & $\bm{\times}$&40.7 ± 2.3 & 25.6 ± 0.7 & \textbf{11.9 ± 0.1} & 2,532 ± 93\\ 
Self Distillation & $\bm{\times}$&66.5 ± 1.6 & 53.6 ± 5.3 & 13.1 ± 0.4& 2,321 ± 103 \\ 
RL Warm-Up & $\bm{\times}$& 68.3 ± 0.7  &55.3 ± 3.8 & 16.0 ± 0.3 & 2,556 ± 62\\ 
SFT & \checkmark & \textbf{72.0 ± 1.5}  & \textbf{60.7 ± 1.5}& 12.1 ± 0.2 & \textbf{1,475 ± 127}\\ 
\bottomrule
\end{tabular}
}
\end{center}
\end{table}

The comparative results in Table~\ref{tbl:initialization} demonstrate that \name{}, when initialized with SFT on the expert experience dataset, significantly outperforms other approaches. This underscores the crucial role of expert trajectories in guiding the agent model towards state-of-the-art performance while maintaining high efficiency. Among the methods without extra data, RL Warm-up shows better performance compared to Self Distillation, while training directly from the Base Model yields notably poor results. These findings further indicate the base model's limited capability in skill library usage and its inadequacy for generating high-quality trajectories during rollouts without proper initialization.

\noindent \textbf{SFT Initialized Baseline GRPO.} 
For a fair comparison, we also conducted experiments using SFT to initialize baseline GRPO. Even with this setting, \name{} demonstrates significantly better performance. Detailed experimental settings, results, and analyses are presented in Appendix~\ref{appendix:sft_grpo}.

\section{Conclusion}
This paper presents a pioneering work in exploring the application of RL to self-improving agents with skill library. To achieve this, we propose a novel RL framework named \name{}, which incorporates GRPO with Sequential Rollout and Skill-integrated Reward. When applied to AppWorld datasets, our approach enables the skill library agent to significantly outperform baselines in both performance and efficiency, paving the way for enhanced self-improvement capabilities with skill libraries through RL.

\section*{Limitations}
In this paper, we conducted experiments exclusively on the AppWorld dataset to demonstrate the effectiveness of \name{} for skill library agent. We chose this dataset because its simulated environment more closely resembles real-world application scenarios. However, we acknowledge that different scenarios may require different agent designs, even when applying similar skill library approaches. In future work, we plan to extend our evaluation to other tool-using agent datasets.

\bibliography{custom}

\clearpage

\appendix

\section{Interaction Format Example of Skill Library Agent} \label{appendix:format}

In this section, we provide examples of the interaction format for Skill Library Agent compared to the baseline agent on AppWorld. The following Figure~\ref{fig:a2} and Figure~\ref{fig:a1} present one interaction step of generating code to login the app Spotify for baseline agent and skill library agent respectively.

\begin{figure}[h]
    \centering
    \includegraphics[width=0.5\textwidth]{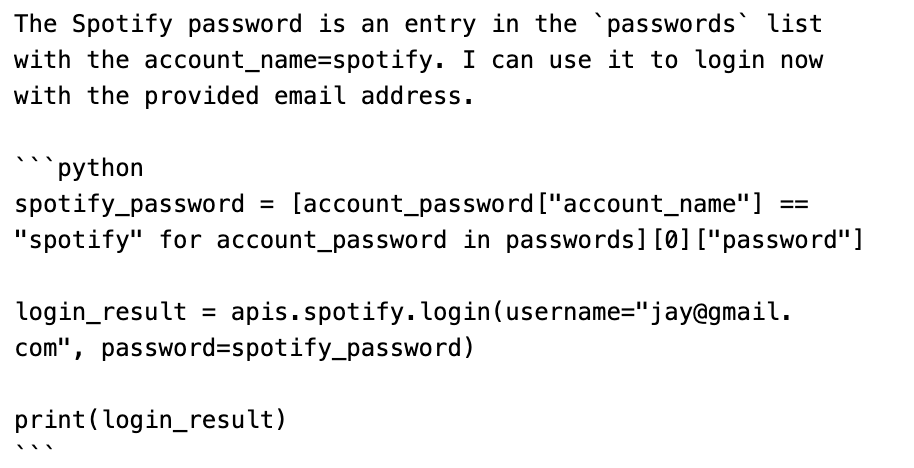}
    \caption{The baseline agent directly generates codes to process the task.}
    \label{fig:a2}
\end{figure}

\begin{figure}[h]
    \centering
    \includegraphics[width=0.5\textwidth]{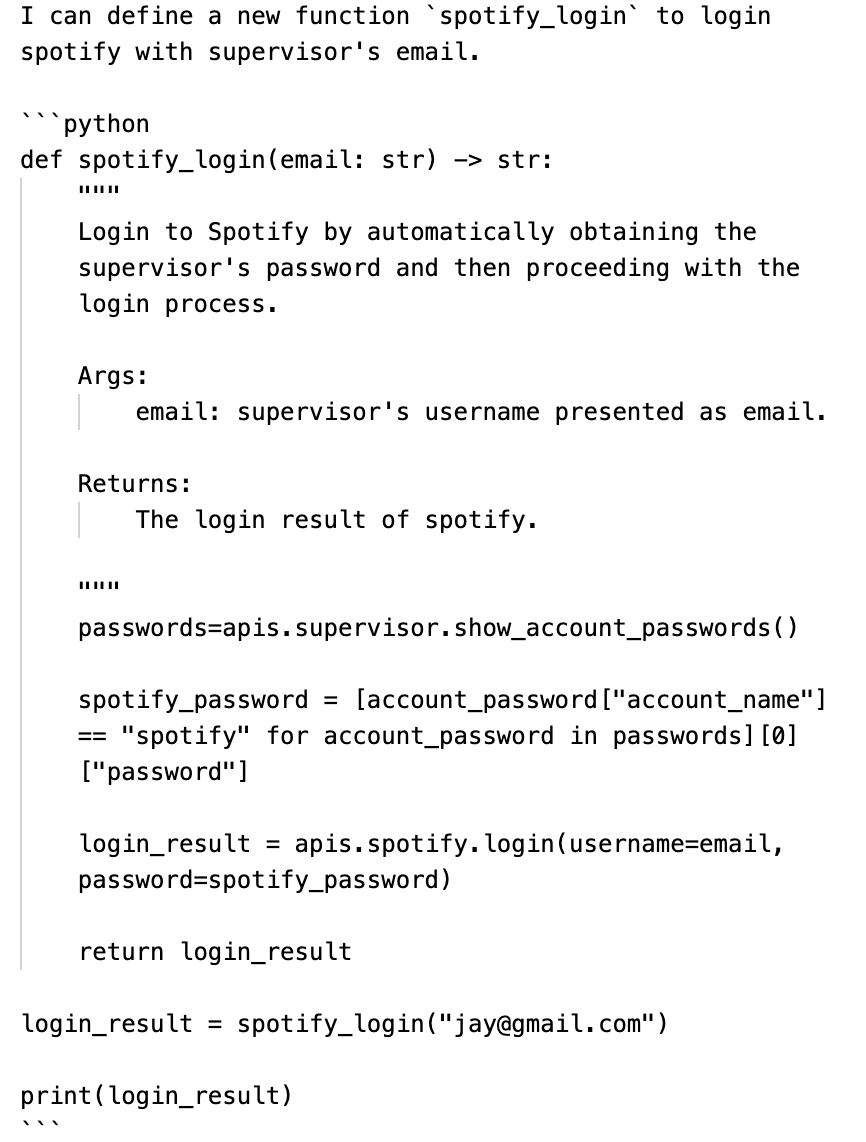}
    \caption{Skill Library Agent will first define a function and then call it to process the task.}
    \label{fig:a1}
\end{figure}

\section{\name{} with Longer Task Chain in Sequential Rollout} \label{appendix:long_chain}

To help readers better understand our \name{} with Sequential Rollout and Skill-integrated Reward, we present a simple but effective case in Section 3.2 where only two example are involved for the task chain. In fact, \name{} has a general format which can be adapted to more than two tasks in the task chain of Sequential Rollout. 

Assuming that we sample $K$ examples from the dataset to form the task chain, denoted as $(q^1,q^2,..,q^K) \sim Q$, the Sequential Rollout process is performed sequentially from $q^1$ to $q^K$ with a skill library, where skills are accumulated into the library and become available to use for all subsequent tasks within the task chain. 

With the Sequential Rollout, here we present the general format for our Skill-integrated Reward. Let $r^1, r^2,...,r^K \in [0,1]$ be the verifiable outcome-based rewards of each task in the long task chain $(q^1,q^2,...,q^K)$, the general format for the Skill-integrated Reward of the $k^\text{th}$ task $R^k$ ($1<k<K$) can be formulated as:
\begin{align*}
R^k & = r^k \\ \nonumber
&+ \mathbf{1}[r^k = 1] * \max_{k<i\leq K}(\mathbf{1}[r^i = 1] * \mathbf{1}_{skill}(q^i|q^k)) \\ \nonumber
&+ \mathbf{1}[r^k = 1] * \mathbf{1}_{skill}(q^{k}|q^1,...,q^{k-1}) \nonumber
\end{align*}
\noindent where $\mathbf{1}_{skill}(q^{k}|q^1,...,q^{k-1})$ evaluates whether the task $q^k$ uses the skills generated from previous tasks $q^1,..q^{k-1}$ in the task chain. 
For the special cases of $k=1$ and $k=K$, the reward becomes slightly different. The first example ($k=1$) begins with an empty skill library, thus cannot utilize previous skills. The last example ($k=K$) cannot have its generated skills verified by subsequent tasks. Therefore, we formulate these boundary cases as:

{\small
\begin{align*}
R^1 & = r^1 + \mathbf{1}[r^1 = 1] * \max_{1<i\leq K}(\mathbf{1}[r^i = 1] * \mathbf{1}_{skill}(q^i|q^1)) \\
R^K & = r^K + \mathbf{1}[r^K = 1] * \mathbf{1}_{skill}(q^{K}|q^1,...,q^{K-1})
\end{align*}
}

To summarize, the basic idea of Skill-integrated Reward design is that beyond the task completion rewards, the agent model receives additional 1.0 skill relevant rewards in two scenarios: 

Skill Generation Reward - when successfully completed task generates skills that are utilized in at least one subsequent task within the task chain and lead to its success task completion;

Skill Usage Reward - when the agent effectively applies previously acquired skills from earlier tasks to successfully complete the current task.

Similar to Section 3.2.4, we describe the general format of \name{} as follows. Given the current policy $\pi_\theta$ and the old policy $\pi_{\theta_{old}}$, we first sample task chain with $K$ examples from the original data distribution: $(q^1, q^2,...,q^K) \sim Q$. For each task pair, we then perform $G$ groups rollout $\{\tau_i\}_{i=1}^G \sim \pi_{\theta_\text{old}}(\cdot|(q^1,q^2,...,q^K))$, where $\tau_i$ represents the Sequential Rollout trajectory collected by sequentially processing $(q^1,q^2,...,q^K)$. To differentiate outputs for each task query in $\tau_i$, we define $o_i^k \in \tau_i$ as the $i^\text{th}$ outputs for $q^k$ in the group, where $k$ is the chain index. The current policy is then optimized by maximizing the following objective function for skill library integrated GRPO:

\begin{align}
\mathcal{J}_{Agent}(\theta) &= \mathbb{E}_{(q^1,...,q^K) \sim Q, \{\tau_i\}_{i=1}^G \sim \pi_{\theta_\text{old}}(\cdot|(q^1,...,q^K))} \nonumber \\
&\frac{1}{G}\sum_{i=1}^G \sum_{k=1}^K \frac{1}{|o_i^k|}\sum_{t=1}^{|o_i^k|}\{\min[s_{i,t}^k\hat{A}_{i}^k, \nonumber \\
&\text{clip}\left(s_{i,t}^k,1-\epsilon, 1+\epsilon\right)\hat{A}_{i}^k]\} \nonumber
\end{align}

\noindent where $s_{i,t}^k = \frac{\pi_\theta(o_{i,t}^k|q^k,\mathcal{M}_{i}^k,o_{i,<t}^k)}{\pi_{\theta_{old}}(o_{i,t}^k|q^k,\mathcal{M}_{i}^k,o_{i,<t}^k))}$; and $\hat{A}_{i}^k=R_{i}^k-\text{mean}(\{R_{i}^k|i=1,2,..,G\})$. $s_{i,t}^k$ represents the importance sampling term; $\hat{A}_{i}^k$ is the advantage computed by the Skill-integrated Reward $R_{i}^k$; $\mathcal{M}_{i}^k$ denotes the skill library integrated for the query $q_i^k$ for group $i$.

To better understand \name{}'s performance with longer task chains during Sequential Rollout, we extended our experiments by combining all three tasks within one scenario into a continuous task chain. This experiment maintained similar settings to the original one described in Appendix~\ref{appendix:sgrpo}, with one key modification: both batch size and mini-batch size were increased to 576 to accommodate the additional rollouts required by the longer task chains. Table~\ref{tbl:longer_chain} compares the performance between \name{} using three-example task chain and the original two-example task chain.

\begin{table}[ht]
\caption{Performance of \name{} with longer task chain.}
\label{tbl:longer_chain}
\begin{center}
\resizebox{\columnwidth}{!}{
\begin{tabular}{ccccc}
\toprule
Task Chain & \multicolumn{4}{c}{Test Normal}  \\
  Length & TGC & SGC& Avg. Steps & Avg. Tokens \\
\midrule
2 & \textbf{72.0 ± 1.5} & \textbf{60.7 ± 1.5} & \textbf{12.1 ± 0.2} & \textbf{1,475 ± 127} \\ 
3 &  70.6 ± 3.6 & 54.8 ± 4.2 & 14.3 ± 0.1 &2,585 ± 13 \\
\bottomrule
\end{tabular}
}
\end{center}
\end{table}

The results indicate that longer task chains do not necessarily improve the skill library agent's performance. This may be attributed to two factors: (1) Reward Distribution Imbalance: Examples in the task chain are typically clustered by similar instructions, leading to most relevant skills being generated during the first example's execution. Consequently, the first example predominantly receives rewards for skill generation, while subsequent examples mainly receive rewards for skill utilization, creating an asymmetric reward structure. (2) Gradient Variance: As the chain lengthens, later tasks begin with increasingly divergent skill libraries, deviating from standard GRPO settings. This deviation potentially leads to larger gradient variance, which may adversely affect the final performance. Besides, the iterative nature of Sequential Rollout process results in substantially increased computational costs as the number of tasks grows. Given these challenges and the empirically suboptimal performance, we ultimately chose a two-example task chain structure. This configuration allows the first example to focus on skill generation while the second emphasizes skill usage.

\section{AppWorld Dataset}\label{appendix:dataset}
In AppWorld dataset \citep{trivedi2024appworld}, autonomous agents are required to complete everyday digital tasks (e.g., sending messages and transfering money to roommates) by consulting API documentation and executing API calls through generated code. Unlike other API usage datasets, AppWorld provides a high-quality execution environment that simulates 9 everyday apps (Amazon, Spotify, Venmo, Gmail, Todoist, SimpleNote, Splitwise, FileSystem, Phone, and ApiDocs). The environment includes totally 457 APIs and features interactions with over 100 simulated users.

To benchmark the agent model performance, AppWorld provides a suite of 750 natural, diverse, and challenging tasks, divided into four splits: Train (105), Dev (60), Test-Normal (168), and Test-Challenge (417). The Test-Challenge set not only contains more test examples but also requires APIs from applications (Amazon and Gmail) that are absent from the Train, Dev, and Test-Normal sets. This design specifically evaluates the models' ability to generalize to unfamiliar APIs.

For accurate evaluation, each task incorporates a manually written program that evaluates the final environment state against predefined criteria, producing a completion rate (0-1) that serves as our outcome-based reward. The stage-based tests accommodate multiple solution paths, enabling more precise reward calculations.
Notably, AppWorld presents a significant challenge even for state-of-the-art LLMs. According to \citet{trivedi2024appworld}, even GPT-4O achieves only about 49\% success rate on Test-Normal tasks and approximately 30\% on Test-Challenge tasks.

Furthermore, the 750 tasks comes from 250 task scenarios, each consisting of three tasks sharing similar instructions under different simulated users. The scenario-based structure of AppWorld makes it well-suited for Sequential Rollout, as tasks within the same scenario naturally form a task chain.

\section{Prompt for Skill Library Agent} \label{appendix:prompt}
Figure~\ref{fig:prompt} illustrates the prompt template used for our skill library agent. Within the prompt, placeholders (marked in red and enclosed in brackets) are designed to be replaced with corresponding content. The template requires an example task to serve as an in-context example, while the skill library section would be replaced with available function skills. All remaining information is derived directly from the AppWorld dataset's task specifications.

\section{Training Details of Baseline GRPO} \label{appendix:grpo}
We built our training framework based on the verl repository \citep{sheng2024hybridflow}, which uses vLLM \citep{kwon2023efficient} for rollout and FSDP \citep{zhao2023pytorch} for gradient update. While our training configuration largely followed \citet{chen2025reinforcement}, there are still some key modifications: we implemented strictly on-policy reinforcement learning and employed full-parameter fine-tuning instead of LoRA. The implementation details of the baseline GRPO used in our paper are shown as follows:

\noindent \textbf{Training Framework.} Our baseline GRPO differs from the original algorithm presented in Section 3.2.1. Here we eliminated the KL penalty and recomputed the advantage of each group as $\hat{A}_{i,t}=r_i-\text{mean}(\mathbf{r})$ without normalized by the standard error.

\noindent \textbf{Dataset.} We only applied the difficulty-1 and difficulty-2 tasks in the Train split of AppWorld dataset for our baseline training.

\noindent \textbf{Rollout Settings.} For each step of training, we performed the rollout process with 1.0 temperature for 36 random sampled examples with $G=8$ agents for each group, resulting in totally 288 rollout. During the agents interact with AppWorld environment, we allowed 40 maximum interaction turns with a limitation of 1,500 output tokens for each turn. Due to the GPU memory limitation, we set the context limitation as 28,048 during rollout. For each task, once its trajectory length tokens exceed the limitation during interaction, the rollout will be stopped and marked as completed. Besides, we observed that the code execution outputs returned by the AppWorld environment may be extremely long (e.g., print all examples in the list). Thus, we applied truncation for returned environment outputs exceeding 12,000 characters and append prompt ``Observation truncated for display.'' afterwards as a truncation notification. After each environment output, we also appended a reminder about the task related information before agents generate the next step. Last but not least, to improve the rollout efficiency, we make an early stop for the entire rollout process when at least 25 interaction steps for unfinished rollout, at least 6 rollouts for each task and 90\% of the total number of rollouts have been collected.

\noindent \textbf{Training Settings.} We performed the whole RL training process on 4xNVIDIA H100 8-GPU nodes. Due to the long-horizon trajectories, we utilized the ulysses sequence parallel \citep{jacobs2023deepspeed}, allocating each example on 4 GPUs. We kept the mini batch size the same as the 288 batch size of rollout. A constant learning rate of 1e-6 and clip the gradient norm to 1 were applied for policy gradient update.

\section{Expert Experience Dataset Generation} \label{appendix:sft}
To collect the expert experience dataset for SFT, we implemented a rejection sampling process under our skill library agent with the advanced model Claude Sonnet 3.5 V2 as the expert. Specifically, we conducted rejection sampling on the training set using the expert model with temperatures ranging from 0.05 to 1.0, with increments of 0.05 across 20 steps. The AppWorld dataset comprises 30 distinct scenarios, and we sequentially deployed the skill library agent through all three tasks within each scenario, allowing skills generated in previous tasks to be directly utilized for subsequent tasks within the same scenario. When processing each task, we implemented a maximum retry limit of 10 attempts, terminating sampling within the whole scenario when this limit is reached. Finally, we retained only those scenarios where either all three tasks or at least the first two tasks are successfully completed, as failures in the second task typically indicate potential issues with the skill generation process. This process finally yielded 1,129 valid examples. 

Using these collected examples as an expert experience dataset, we employed Llama-Factory \citep{zheng2024llamafactory} to perform full parameter fine-tuning with 4xNVIDIA H100 8-GPU nodes. During fine-tuning, since all trajectories in the expert experience dataset consist of multiple turn interactions, we did gradient update exclusively on the agent responses with prompt and environment outputs masked. The model was trained with a batch size of 128 and employed a learning rate of 1e-6, using a cosine scheduling strategy with 0.1 warmup ratio. Due to Llama-Factory's limitations in sequence parallel, we set the maximum token length to 12,500, and any examples exceeding this limit were excluded from the dataset.



\section{Training Details of \name{}} \label{appendix:sgrpo}
To ensure a fair comparison with the baseline GRPO method, we maintained consistency in most training parameters. However, the introduction of the Sequential Rollout process necessitated certain modifications. Below, we detail the specific adjustments made to our training configuration. 

\noindent \textbf{Training Framework.} Refer to Section 3.2 for \name{} framework used for training.

\noindent \textbf{Dataset.} While we continued to utilize difficulty-1 and difficulty-2 tasks from the Train split of AppWorld as our training dataset, we implemented a revised sampling strategy to accommodate the Sequential Rollout process. We adopt a scenario-based sampling approach to construct the required task chains. Specifically, AppWorld tasks are inherently organized into scenarios, with each scenario comprising three distinct tasks. Our sampling process consisted of two steps: we first selected a certain number of task scenarios, and then within each selected scenario, we sampled two tasks to form a task chain.

\noindent \textbf{Rollout Settings.} Because the Sequential Rollout process requires iterative interactions within the task chain, the rollout completion time approximately doubles compared to the baseline if we maintain the same total number of rollouts. To accelerate the training process, we required a larger number of total rollouts. For each step of training, we decided to use all scenarios in the Train split of AppWorld with difficulty-1 and difficulty-2 tasks, which only contains 24 scenarios. Sampling two task examples from those scenarios resulted in 48 tasks. Under the same agent number $G=8$ for each group, we finally obtained 384 rollouts for each step.

\noindent \textbf{Training Settings.} The only change during training was the mini batch size for policy gradient update. Here we aligned it to the batch size of rollout, which is 384.

\begin{figure}[h]
    \centering
    \includegraphics[width=0.5\textwidth]{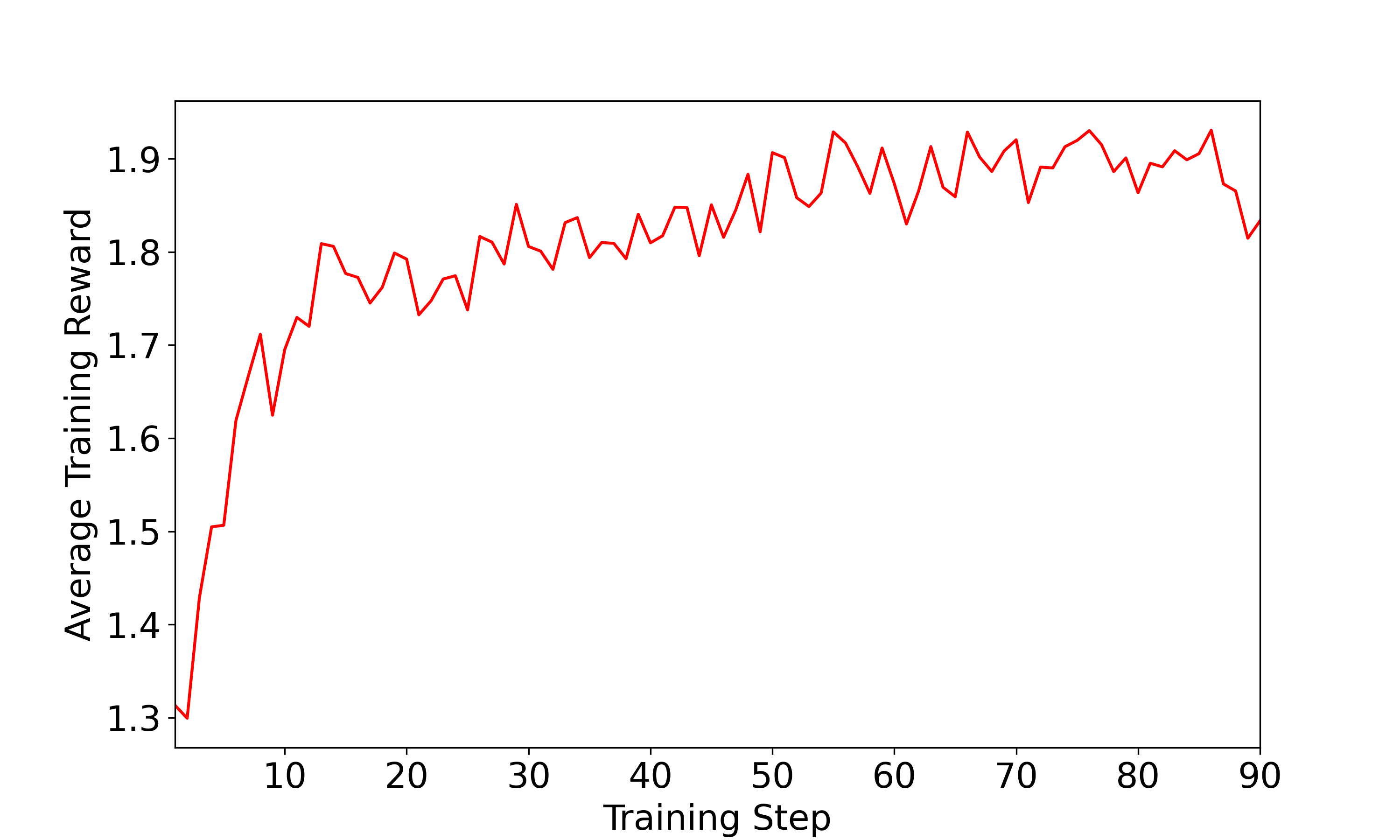}
    \caption{Training curve of \name{}.}
    \label{fig:train}
\end{figure}

\begin{figure}[h]
    \centering
    \includegraphics[width=0.5\textwidth]{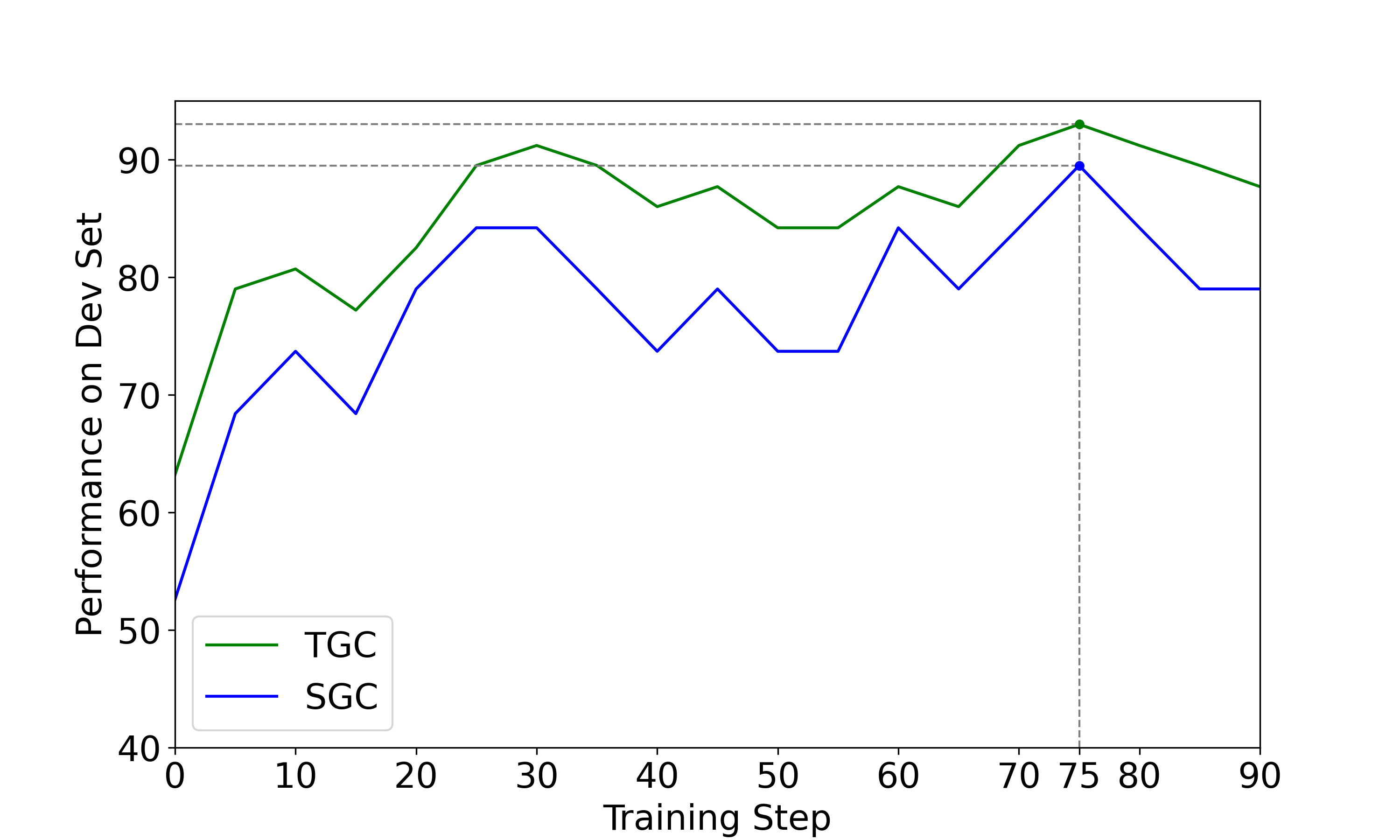}
    \caption{SGC and TGC scores on Dev set for each 5 training steps.}
    \label{fig:eval}
\end{figure}

\noindent \textbf{Training Details.} We present the training curve with average training reward per step in Figure~\ref{fig:train}. During training, we saved model checkpoints every 5 steps and evaluated their performance on the Dev set. The checkpoint achieving the highest combined TGC and SGC scores on the Dev set was selected as our final model. Evaluation curve showing the TGC and SGC scores on the Dev set is illustrated in Figure~\ref{fig:eval}. The figure clearly demonstrates that the model achieves optimal performance in both TGC and SGC metrics at step 75. Thus, we selected the model checkpoint at step 75 as our final agent model.

\section{Task Execution Examples with Different Models} \label{appendix:example}

\begin{figure}[h]
    \centering
    \includegraphics[width=0.5\textwidth]{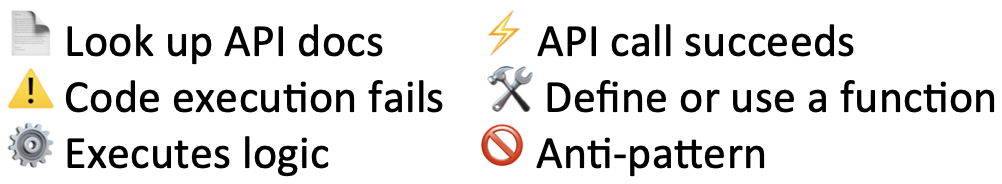}
    \caption{Action Explanations. Code execution failures encompass various issues including unsuccessful API calls, errors in skill generation, and improper skill usage. Anti-patterns primarily refer to the key actions that ultimately lead to task failures.}
    \label{fig:emoji}
\end{figure}

This section presents several real task execution examples on AppWorld, comparing agent models from the baseline GRPO and each stage of our approach. These examples demonstrate the advantages of the skill library agent and the improvements in skill library usage achieved by \name{}. Due to the extensive length of the trajectories, all examples are presented in a summarized format. For each interaction turn, we represent the agent's action using an emoji accompanied with a brief description. Figure~\ref{fig:emoji} provides explanations for the meaning of each emoji.

\begin{figure*}[ht]
    \centering
    \includegraphics[width=1.0\textwidth]{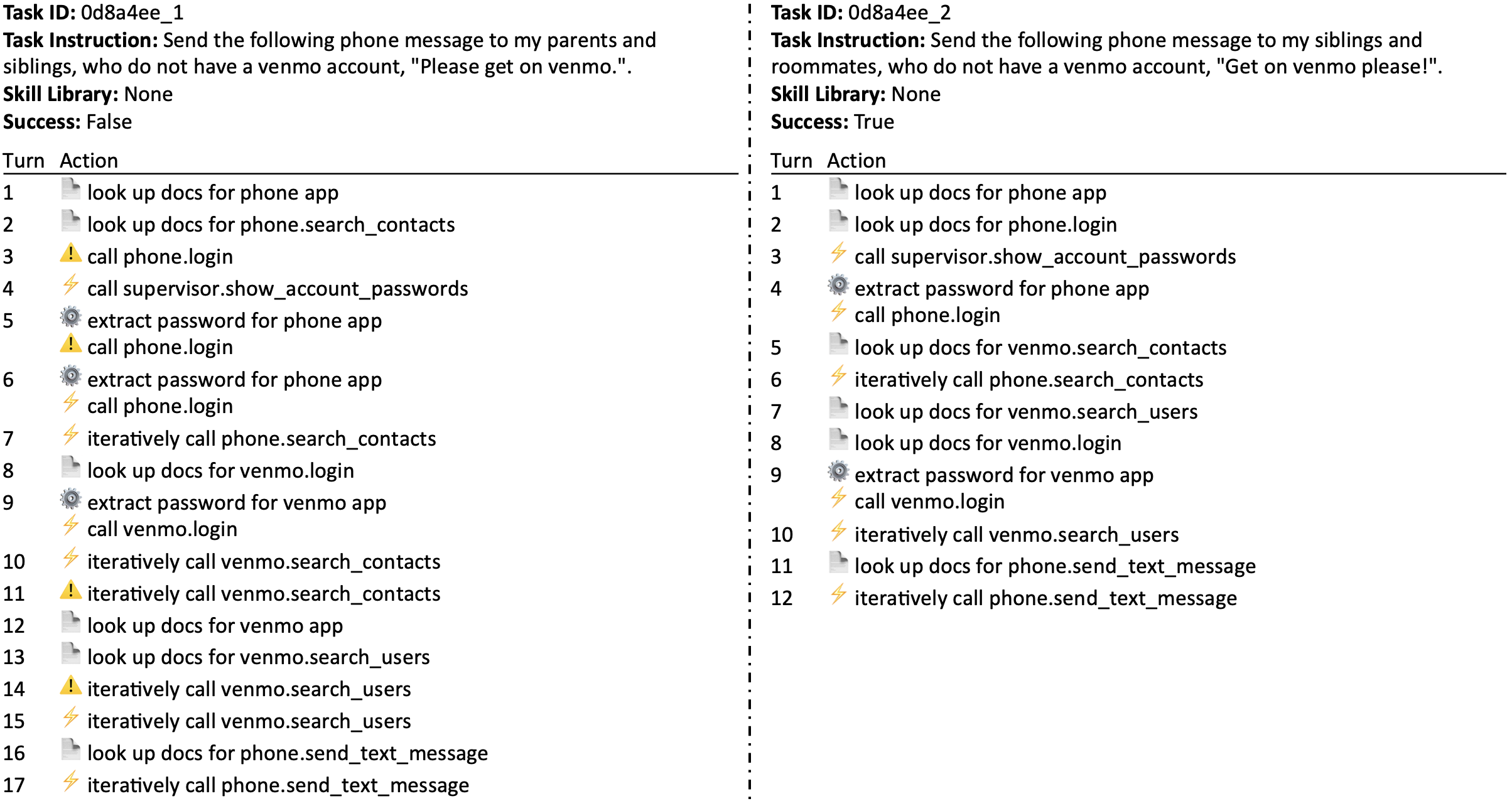}
    \caption{Task Execution Examples for Baseline GRPO}
    \label{fig:example_grpo}
\end{figure*}

\begin{figure*}[ht]
    \centering
    \includegraphics[width=1.0\textwidth]{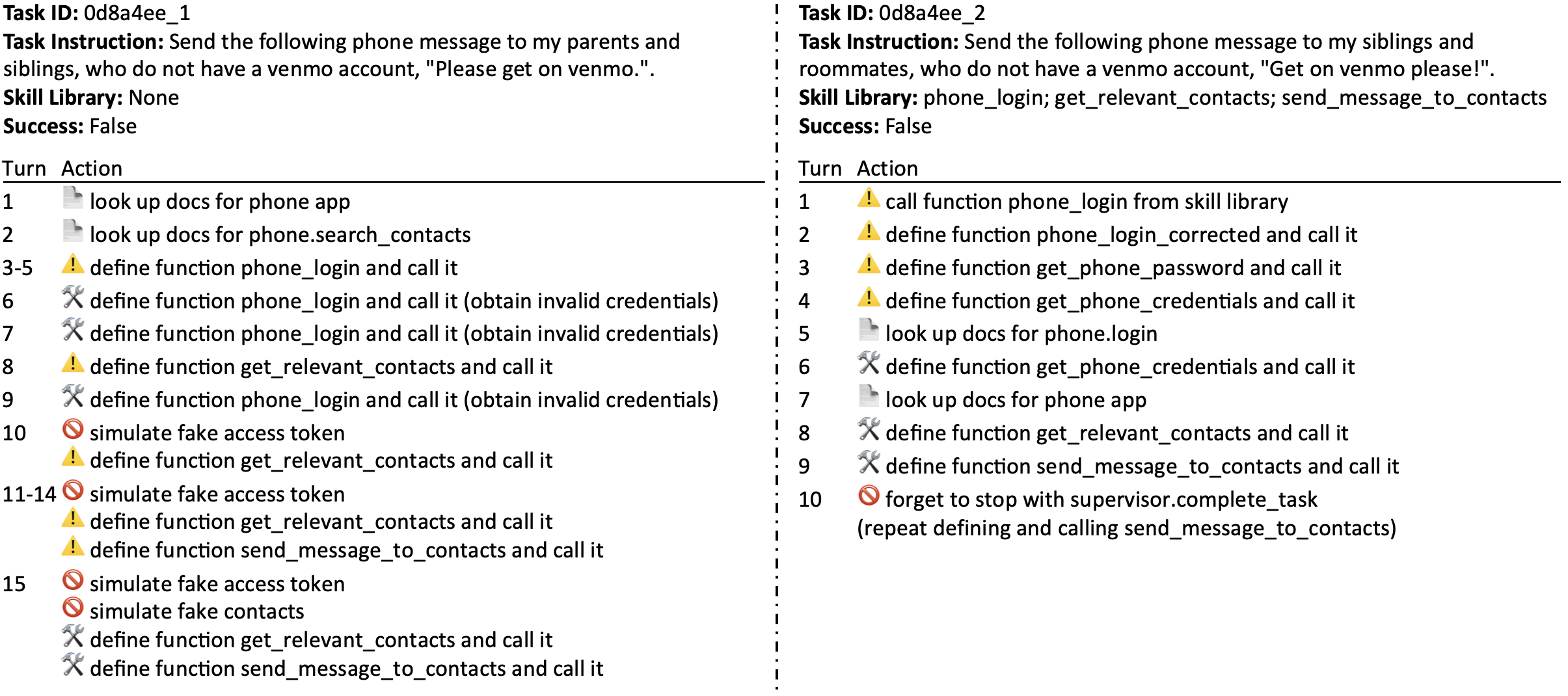}
    \caption{Task Execution Examples for Skill Library Agent}
    \label{fig:example_slagent}
\end{figure*}

\begin{figure*}[ht]
    \centering
    \includegraphics[width=1.0\textwidth]{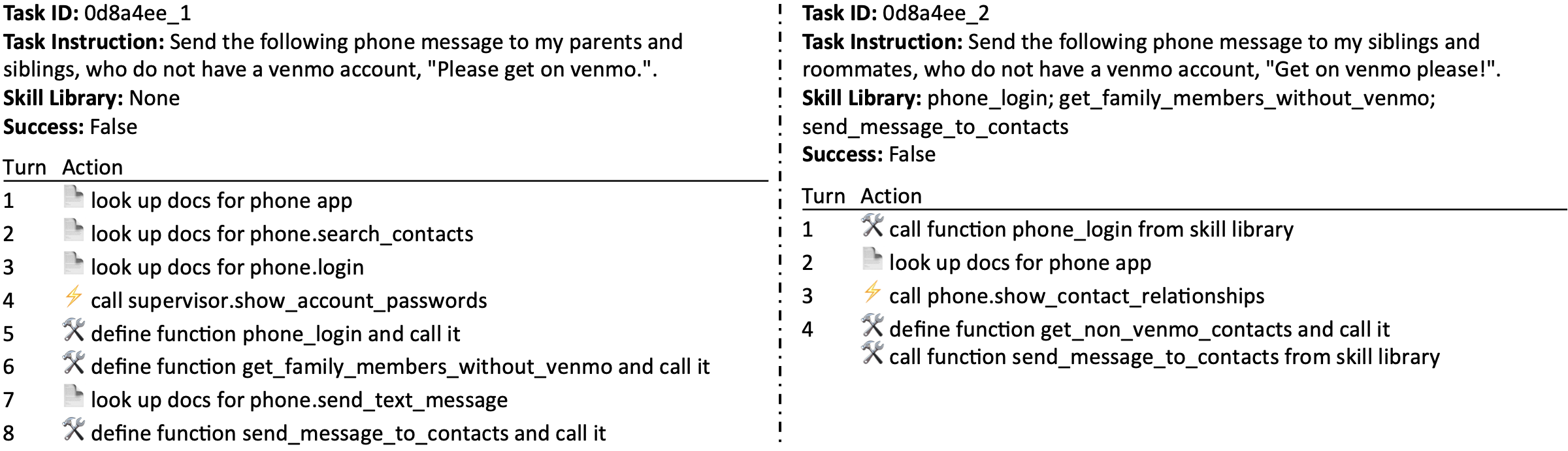}
    \caption{Task Execution Examples for SFT}
    \label{fig:example_sft}
\end{figure*}

\begin{figure*}[ht]
    \centering
    \includegraphics[width=1.0\textwidth]{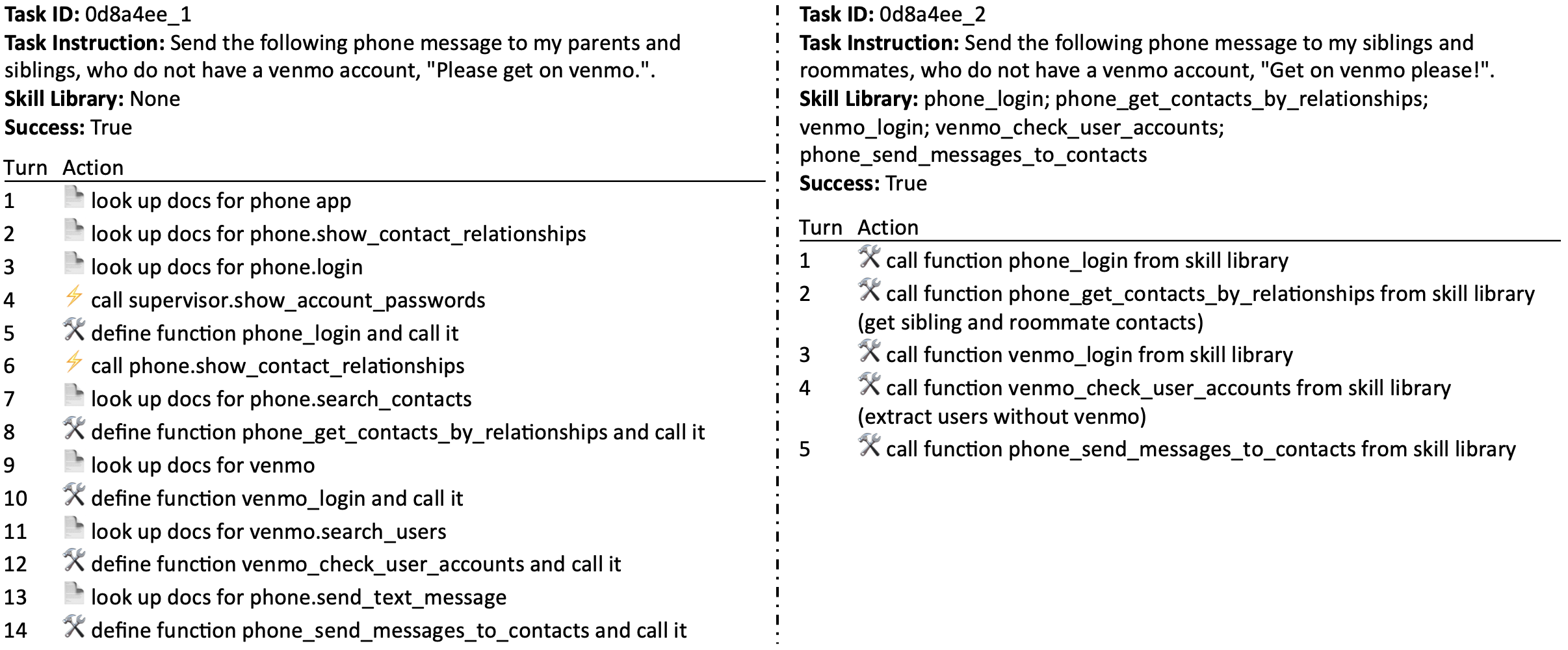}
    \caption{Task Execution Examples for \name{}}
    \label{fig:example_sage}
\end{figure*}

To clearly demonstrate the skill generation and usage process, we present two task execution examples for each agent model under the same scenario. Examples of the baseline GRPO are shown in Figure~\ref{fig:example_grpo}. The results for the Skill Library Agent, SFT, and \name{} stages of our approach are presented in Figures~\ref{fig:example_slagent}, \ref{fig:example_sft}, and \ref{fig:example_sage} respectively.

Analyzing the task execution examples across different stages of our approach reveals several key observations. The Skill Library Agent, before training, demonstrates poor performance in task execution with the skill library. It often requires multiple attempts to define a executable function, and even then, errors may still occur. Additionally, it tends to simulate data for task processing and exhibits unexpected behaviors such as repetitive generation. 

After SFT, the agent model shows significantly reduced interaction steps and fewer errors in skill generation and usage. However, while these patterns learned from expert experience data minimize basic errors, the SFT model still struggles to achieve successful task completion.

In comparison to the baseline GRPO, agent model trained by \name{} substantially improves task efficiency by eliminating the need for step-by-step API calls and reducing redundant actions in similar tasks.

\section{Retrieval Method Ablation Study} \label{appendix:retrieval}
In this section, we introduce each retrieval method presented in Table~\ref{table4} and provide more analysis for the results.

\subsection{Details of Retrieval Methods}
\noindent \textbf{Same Scenario.} The Same Scenario method serves as our paper's default evaluation setting, effectively utilizing the scenario labels provided in the dataset. Under this approach, we construct a task chain comprising three tasks within each scenario, combined with a skill library. The agent performs these tasks sequentially, with skills generated from the first two examples being accumulated in the skill library. This accumulated knowledge is then made available for the subsequent two examples. Importantly, each skill library is constrained to its specific task scenario.

\noindent \textbf{Query N-gram.} The core concept of this retrieval method is to leverage skills generated from tasks with similar queries~\cite{xu2025survey}. Under this approach, each query is associated with its own skill library, which can be directly inherited once being retrieved. For simplicity, we initially implement a model-free retrieval approach based on 2-gram Jaccard Similarity to identify the most similar query. To ensure retrieval quality, we establish a threshold of 0.5 for the Jaccard Similarity score; any retrieved queries falling below this threshold are discarded.

\noindent \textbf{Query Embedding.} This retrieval method shares the same fundamental principle as Query N-gram, but employs a more advanced text-embedding model to compute embedding similarity for query retrieval. Specifically, we utilize the all-MiniLM-L6-v2 model \citep{reimers-2019-sentence-bert} for computing query embeddings with a threshold of 0.65 for the cosine similarity.

\noindent \textbf{Skill Embedding.} Skill Embedding retrieves relevant skills from the skill library for each task query using a standard retrieval model. Each newly generated skill is encoded into embeddings and indexed in the skill library for future retrieval purposes. We employ Qwen3-Embedding-0.6B \citep{qwen3embedding} as our retrieval model and keep the top 5 retrieved skills for usage. This model differs from the general text-embedding model used for Query Embedding because it is specifically trained for document retrieval, where we treat skills as documents and task instructions as queries.

\subsection{Further Analysis}
Among the three retrieval methods studied in our ablation study, Query N-gram achieves performance most similar to the ideal Same Scenario case. This is because tasks under the same scenario in AppWorld share the same query structure. Some tasks within the same scenario even use identical query under different simulated users. Queries within the same scenario naturally share high similarity under N-gram metric. With a proper threshold, the skill library can be effectively constrained to skills within the same scenario.

Query Embedding shows slightly lower performance in TGC and SGC but higher efficiency with fewer Avg. Steps and Tokens. Since text embedding models primarily focus on semantic meaning rather than structural patterns, and queries with different structures can convey similar semantic meanings, it is challenging for Query Embedding restricting queries to a single scenario. 
On the other hand, with Query Embedding, most initial examples within each task scenario can retrieve similar queries from other scenarios. While this broader skill usage leads to fewer interaction steps and tokens, it doesn't necessarily improve accuracy due to difficult cross-scenario skill adaptation.

The Skill Embedding method demonstrates lower performance compared to other approaches. This may be attributed to the inherent difficulty in retrieving useful skill functions based on queries, as it requires understanding tool usage rather than just text relevance. Future work could explore the development of skill/tool-specific retrievers to enhance skill library agent performance.

\section{Reward Design Ablation Study} \label{appendix:reward}
This section supplements the reward design ablation study by providing detailed descriptions of alternative reward designs and additional analysis of skill library usage.

\subsection{Details of Reward Designs}\label{appendix:reward1}

\noindent \textbf{Outcome-based Reward.} Outcome-based Reward design solely relies on the task completion rate as the reward. In this case, skill learning is driven exclusively by Sequential Rollout.

Formally, let $ r^1, r^2 \in [0,1]$ denote the verifiable task completion rate of the task chain $(q^1, q^2)$ collected from Sequential Rollout. The Outcome-based Rewards can be directly formulated as $R^1_O=r^1$ and $R^2_O=r^2$. Similar to the Skill-integrated Reward design, we impose a -1.0 penalty reward specifically on responses where the agent provides no code and terminates the task.

\noindent \textbf{Chain-based Reward.} Chain-based Reward design adds a bonus reward of 1.0 when all tasks within the task chain are successfully completed, rather than rewarding the skill generation and usage throughout the execution. This design aims to test the necessity of tracking precise skill usage during the rollout process.

Formally, let $ r^1, r^2 \in [0,1]$ denote the verifiable outcome-based rewards of the task chain $(q^1, q^2)$ collected from Sequential Rollout. We can formulate the Chain-based Rewards $R^1_C$ and $R^2_C$ as:
\begin{equation*}
\begin{cases}
R^1_C =r^1 + \mathbf{1}[r^1 = 1] * \mathbf{1}[r^2 = 1] \\
R^2_C =r^2 + \mathbf{1}[r^1 = 1] * \mathbf{1}[r^2 = 1]
\end{cases}
\end{equation*}
\noindent where indicator $\mathbf{1}[r = 1]$ represents whether the outcome-based reward equals 1.
Similar to the Skill-integrated Reward design, we impose a -1.0 penalty reward specifically on responses where the agent provides no code and terminates the task.

\subsection{Skill Library Usage Analysis}\label{appendix:reward2}

\begin{figure}[ht]
    \centering
    \includegraphics[width=0.50\textwidth]{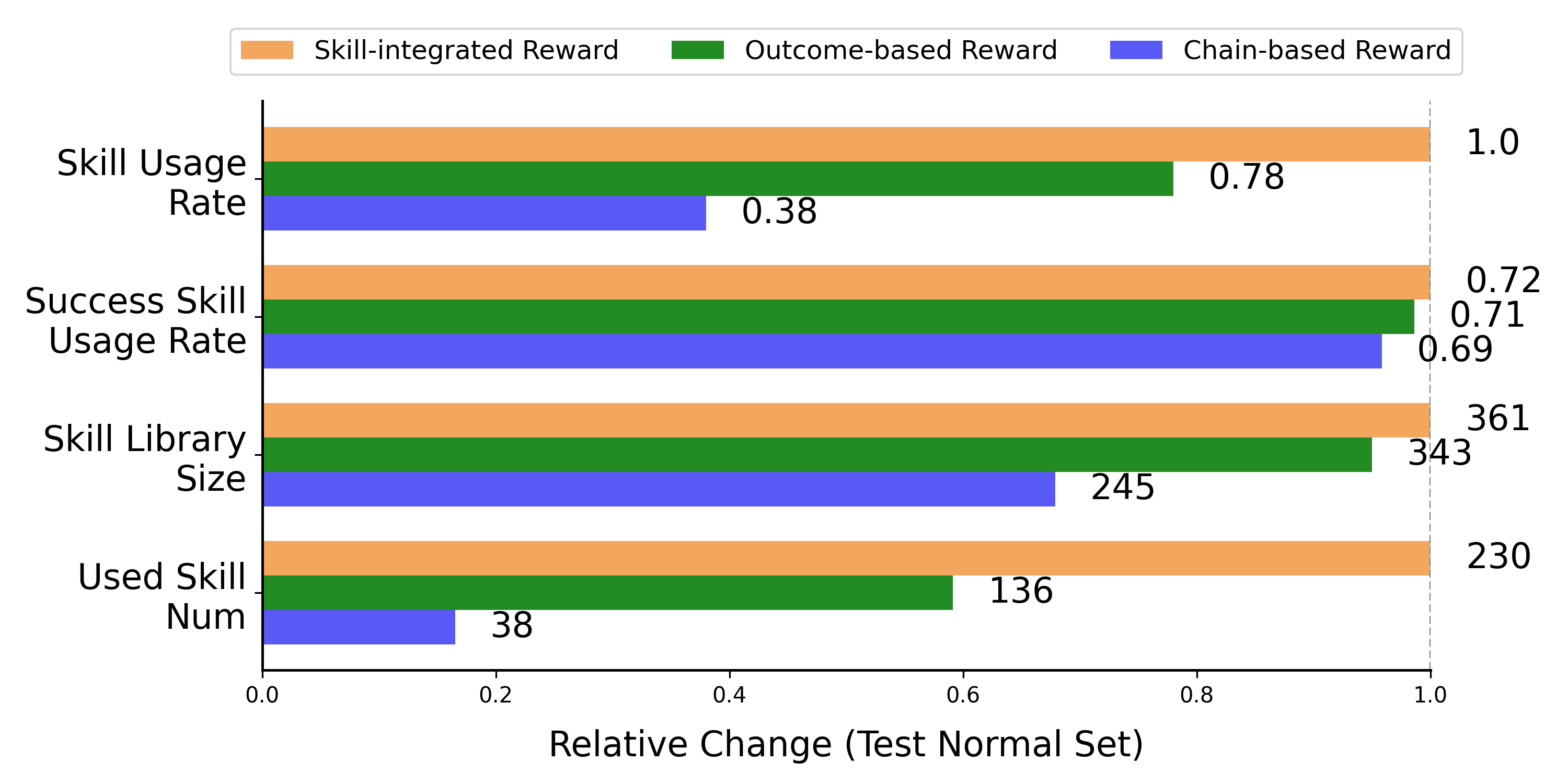}
    \caption{Analysis of skill usage patterns across different reward designs. Performance metrics are shown as ratios relative to Skill-integrated Reward design, with numerical values annotated.}
    \label{fig:ablation_analysis}
\end{figure}

To better illustrate the advantages of Skill-integrated Reward design, we conducted a detailed skill library usage analysis similar to Section 4.3 for agent models trained with different reward designs. Figure~\ref{fig:ablation_analysis} summarizes the relative performance changes when comparing alternative reward designs to our Skill-integrated Reward. 
The results reveal that while the agent model trained with Skill-integrated Reward exhibits significantly more skill usage behaviors (Skill Usage Rate, Skill Library Size and Used Skill Num), all three reward designs maintain similar Success Skill Usage Rate. This indicates that our reward design substantially enhances skill usage frequency, but successful skill application appears to be a general behavior achievable across different reward designs.
Furthermore, the smaller Skill Library sizes and fewer Used Skill Nums in both Outcome-based and Chain-based Rewards explain their higher Avg. Steps but lower Avg. Tokens compared to Skill-integrated Reward.
This pattern emerges because while skill usage typically reduces interaction steps, the generation of skill functions generally requires more tokens. Interestingly, the Chain-based Reward shows notably lower skill usage behaviors compared to even the Outcome-based Reward. This phenomenon might be attributed to the early training stages where, despite successful task completion in the task chain, skill usage rates remain low. Consequently, the additional reward added in Chain-based Reward may inadvertently reinforce behavior patterns that exclude skill usage.

\section{RL Initialization Methods}\label{appendix:init} This section details various initialization methods before implementing \name{}.

\noindent \textbf{Base Model.} In this approach, we directly apply the base model, Qwen2.5-32B-Instruct, to the RL process without any additional initialization.

\noindent \textbf{Self Distillation.} Like SFT initialization, Self Distillation employs supervised fine-tuning. However, instead of using Claude to generate the expert experience dataset, it utilizes the base model (Qwen2.5-32B-Instruct) for data generation.

\noindent \textbf{RL Warm-Up.} This method initializes \name{} through a preliminary RL process. It employs baseline GRPO with the skill library agent prompt, but without actual skill library involvement, to familiarize the model with the agent format before implementing \name{}.

\section{SFT Initialized Baseline GRPO}\label{appendix:sft_grpo}

\begin{figure*}[ht]
    \centering
    \includegraphics[width=1.0\textwidth]{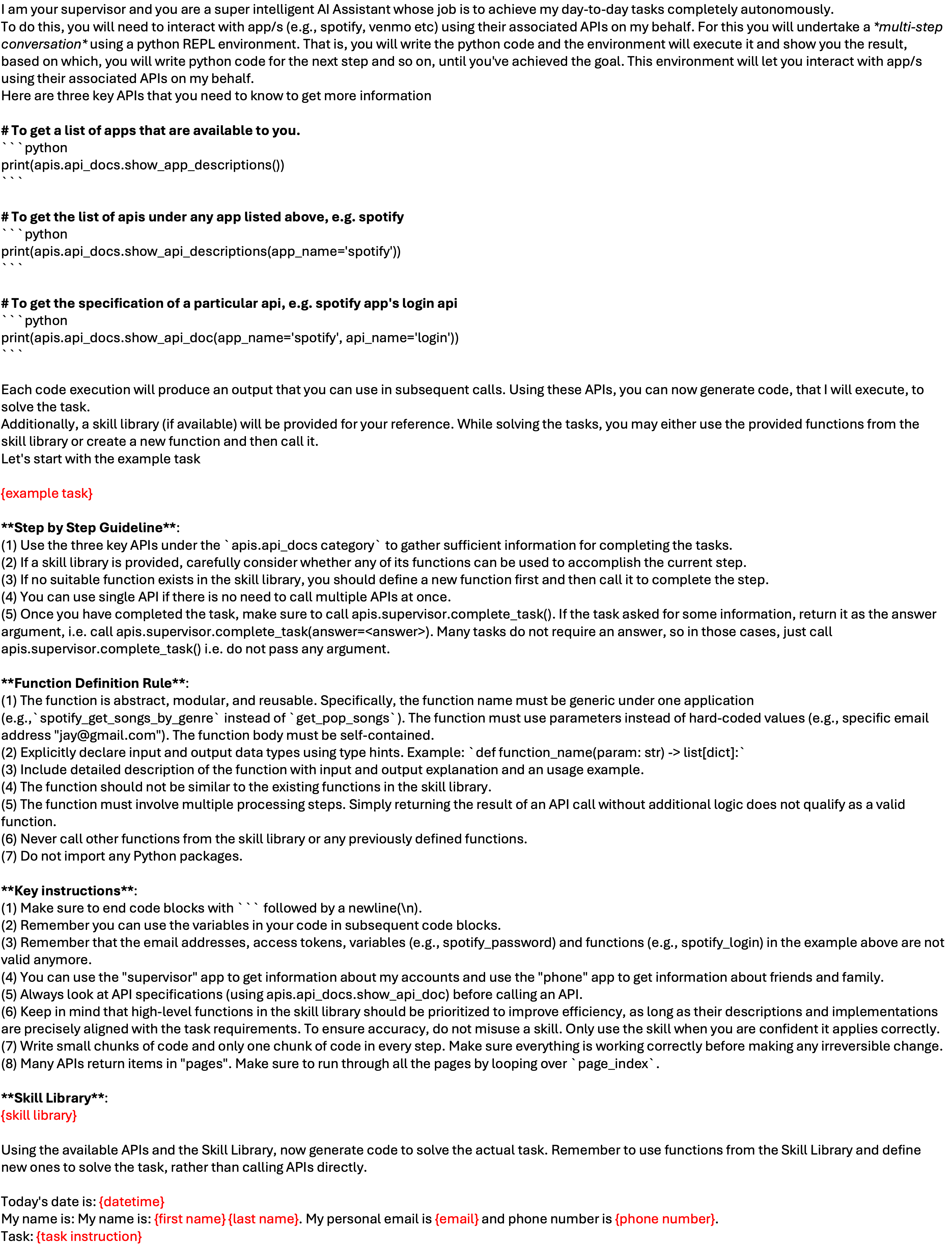}
    \caption{Prompt for Skill Library Agent.}
    \label{fig:prompt}
\end{figure*}

For the SFT initialized baseline GRPO, we used the same SFT model checkpoint as in \name{}. Unlike the original baseline GRPO settings described in Appendix~\ref{appendix:grpo}, this SFT-initialized version utilizes the skill library agent framework without engaging the skill library, rather than using the original ReAct framework. Specifically, we employed the skill library agent prompt illustrated in Figure~\ref{fig:prompt} to guide the rollout process, wherein the model first generates a function and then executes it to process the task. Since the baseline GRPO operates without the skill library, we applied an empty skill library to the prompt placeholder for each task during rollout. The comparative results among this SFT initialized baseline GRPO, the original baseline GRPO, and \name{} are presented in Table~\ref{tbl:sft_grpo}. To align with the training process, only \name{} in Table~\ref{tbl:sft_grpo} performs the evaluation with skill library involved.

From the table, we observe that SFT may not benefit the baseline training, resulting in even worse performance than the original baseline. On the other hand, SFT initialized baseline GRPO demonstrates lower average generated tokens. This phenomenon may be attributed to the agent model remaining constrained by the initial patterns acquired through SFT.

\begin{table}[ht]
\caption{Performance of SFT Initialized Baseline GRPO.}
\label{tbl:sft_grpo}
\begin{center}
\resizebox{\columnwidth}{!}{
\begin{tabular}{ccccc}
\toprule
 \multirow{2}{*}{Method} & \multicolumn{4}{c}{Test Normal}  \\
 & TGC & SGC& Avg. Steps & Avg. Tokens \\
\midrule
GRPO &  69.2 ± 2.7 & 51.8 ± 5.8 & 16.4 ± 0.2& 3,613 ± 200\\ 
SFT Initialized GRPO & 66.1 ± 1.3 & 51.2 ± 0.8 & 12.8 ± 0.1 & \textbf{1,284 ± 18} \\ 
\name{} & \textbf{72.0 ± 1.5} & \textbf{60.7 ± 1.5} & \textbf{12.1 ± 0.2} & 1,475 ± 127 \\ 
\bottomrule
\end{tabular}
}
\end{center}
\end{table}

\end{document}